\pdfoutput=1

\documentclass[11pt]{article}
\usepackage[dvipsnames, table, svgnames, dvipsnames]{xcolor}

\usepackage{ACL2023}

\usepackage{times}
\usepackage{latexsym}

\usepackage[T1]{fontenc}

\usepackage[utf8]{inputenc}

\usepackage{microtype}

\usepackage{inconsolata}

\usepackage{graphicx}
\usepackage{multirow}
\usepackage{booktabs}
\usepackage{arydshln}
\usepackage{colortbl}
\usepackage{pifont}
\newcommand{\xmark}{\ding{55}}%
\usepackage{wrapfig}
\usepackage{enumitem}
\usepackage{makecell}

\usepackage{amsmath,amsfonts,bm}
\usepackage[nameinlink]{cleveref}

\crefformat{section}{\S#2#1#3} 
\crefname{algorithm}{Alg.}{Algs.}
\crefformat{subsection}{\S#2#1#3}
\Crefname{equation}{Eq.}{Eqs.}
\Crefname{figure}{Fig.}{Figs.}

\makeatletter   
\newcommand{\sveryshortarrow}[1][3pt]{\mathrel{%
    \vcenter{\hbox{\rule[-.5\fontdimen8\scriptfont3]
               {\scriptratio\dimexpr#1\relax}{\fontdimen8\scriptfont3}}}%
   \mkern-4mu\hbox{\let\f@size\sf@size\usefont{U}{lasy}{m}{n}\symbol{41}}}}
\makeatother

\def\eqref#1{equation~\ref{#1}}

\def\1{\bm{1}}

\def\m1{{\bm{1}}}

\DeclareMathAlphabet{\mathsfit}{\encodingdefault}{\sfdefault}{m}{sl}
\SetMathAlphabet{\mathsfit}{bold}{\encodingdefault}{\sfdefault}{bx}{n}

\def\gL{{\mathcal{L}}}

\title{PromptSum: Parameter-Efficient Controllable
 Abstractive Summarization}

\author{Mathieu Ravaut$^1$$^,$$^2$,~
Hailin Chen$^1$$^,$$^3$,~
Ruochen Zhao$^1$,~ \\
{\bf Chengwei Qin}$^1$,~
{\bf Shafiq Joty}$^*$$^1$$^,$$^3$~
{\bf Nancy F. Chen}$^2$ \\
$^1$ Nanyang Technological University, Singapore\\
$^2$ Institute of Infocomm Research (I$^{2}$R), A$^{*}$STAR, Singapore\\
$^3$ Salesforce AI
}
\begin{document}

\maketitle
\def\thefootnote{*}\footnotetext{Work done when the author was on leave from NTU.}\def\thefootnote{\arabic{footnote}}

\begin{abstract}

Prompt tuning (PT), a parameter-efficient technique that only tunes the additional prompt embeddings while keeping the backbone pre-trained language model (PLM) frozen, has shown promising results in language understanding tasks, especially in low-resource scenarios. However, effective prompt design methods suitable for generation tasks such as summarization are still lacking. At the same time, summarization guided through instructions (discrete prompts) can achieve a desirable double objective of high quality and controllability in summary generation. Towards a goal of strong summarization performance under the triple conditions of parameter-efficiency, data-efficiency, and controllability, we introduce PromptSum, a method combining PT with a multi-task objective and discrete entity prompts for abstractive summarization. Our model achieves competitive ROUGE results on popular abstractive summarization benchmarks coupled with a strong level of controllability through entities, all while only tuning several orders of magnitude less parameters.\footnote{All code and model checkpoints will be available at: \url{https://github.com/ntunlp/PromptSum}}

\end{abstract}

\section{Introduction}
\label{sec:intro}
Pre-training large language models and adapting them to downstream tasks through fine-tuning has become the dominant paradigm in NLP \citep{raffel2019exploring,radford2019language,lewis2019bart}, including in summarization \citep{zhang2020pegasus}. However, full-model fine-tuning requires storing an entire set of weights for each downstream task, prohibiting simultaneous multi-task inference as the models become larger \citep{chowdhery2022palm,thoppilan2022lamda}. Also, it requires careful considerations to avoid overfitting and forgetting, especially when the dataset is small as in few-shot tasks \citep{kaplan2020scaling}. 


Prompt engineering has recently gained popularity as a low-cost alternative for adapting PLMs to downstream tasks. Such prompts are generally constructed either by finding  suitable discrete task instructions (e.g., ``TL;DR'' for summarization) with possibly a few examples  \citep{brown2020language,schick-schutze-2021-exploiting} or by tuning embeddings of \emph{soft} prompts \citep{li2021prefix,lester2021power}. As a specific conditioning of the PLM, prompt-based adaptation has achieved comparable (or even better) performance to standard fine-tuning in low-resource scenarios, while using very few or even \emph{zero} learnable parameters \citep{brown2020language, li2021prefix, lester2021power}, dramatically cutting down model storage costs when adapting the PLM to a new setup.

Specifically, soft prompt tuning \citep{lester2021power} adapts the nature of discrete prompts to continuous soft prompts whose embeddings are \emph{learned} with backpropagation.  This way, the model is offered greater flexibility and capacity in learning  prompts, as opposed to manually finding an effective discrete prompt. Soft prompts have brought great success in many language understanding tasks \citep{qin2021learning,PPT,SPoT,PT_transfer}, and have also shown promising results in language generation  \citep{li2021prefix, qin2022lfpt}. Despite this, one important drawback of soft prompt tuning is that it lacks the human-explainable nature of discrete text prompts, thus sacrificing  user controllability.

Moreover, for a complex and less constrained generation task like abstractive summarization, guiding the model with additional discrete signals, such as entities or keywords, can significantly enhance the performance as it helps the generative model to be faithful and on topic \citep{dou2020gsum}. An entity chain (sequence of entities) can be viewed as a high-level \emph{content plan} that can bootstrap the generation process \citep{liu2021controllable, puduppully2019data}. Entities are also a strong proxy for topic saliency \citep{Barzilay:2005}. Intuitively, when a model is provided with extra information on the important entities to incorporate into the output summaries, it is expected to be more accurate and controllable, and the summary will more likely cover the important facts. Similar to Neuro-Symbolic approaches \citep{garcez2022neural}, we try to approximate the human reasoning process of plannning content with entities first before drafting the summaries. Controllability is an essential aspect in summarization as it is a subjective task, where several different yet all valid summaries can be generated from the same source, for instance when different aspects are considered \citep{liu2020conditional, ahuja2022aspectnews}.

Recent work in summarization has indeed incorporated high-level content planning or controllability. CTRLSum \citep{he2020ctrlsum} takes a discrete instruction as input to enable control on several summarization aspects such as output entities or summary length. \citet{narayan2021planning,narayan2022well} propose FROST that trains the same model to first generate an entity chain, and then a summary by conditioning on the entity chain in an auto-regressive manner. They also adapt PEGASUS \citep{zhang2020pegasus} pre-training objective to include an entity chain during pre-training. This leads to both higher ROUGE scores and a powerful mechanism to control summary generation, which allows for reducing hallucinations by dropping entities not present in the source. Nevertheless, these methods require tuning the entire PLM, on the entire dataset.

In this work, we combine the best of both worlds. On the one hand, we leverage soft prompts to achieve parameter-efficiency in data-efficient setups. On the other hand, we use discrete prompts to induce controllability in summarization. Doing so, we build \emph{\textbf{the first abstractive summarization model operating under the triple constraint of parameter-efficiency, data efficiency, and controllability}}, all while maintaining high performance compared to a strong PEGASUS baseline. We point out that combining just two out of these three properties had never been attempted before, let alone all three of them. Our model PromptSum is multi-task: it uses a soft prompt to generate the entity chain, then another soft prompt and the generated entity chain for summary generation (\Cref{fig_1}). To ensure high quality in both tasks, we introduce a new multi-task prompt-tuning pre-training mechanism. We demonstrate the strength of our parameter-efficient controllable summarization approach in both data efficient and full data setups on four major summarization benchmarks: CNN/DM  \citep{hermann2015teaching}, XSum \citep{narayan2018don}, BillSum \citep{kornilova2019billsum} and SAMSum \citep{gliwa2019samsum}. Through extensive experiments, we show that PromptSum enables great flexibility since the user can simply input entities to condition on, and we make several important findings:

\begin{itemize}[leftmargin=*]
    \item Our introduced multi-task pre-training is vital to enable parameter-efficient controllable summarization in data efficient setups (\Cref{sec:4_2}), and without it model performance collapses.

    \item Controllability is strong, and the generated entity chain heavily impacts both the entities in the summary and the overall summary quality (\Cref{sec:4_2}, \Cref{sec:4_3}).

    \item By removing hallucinated entities in the generated entity chain, we show how to significantly reduce hallucinations in the summary (\Cref{sec:4_4}).
    
\end{itemize}

\section{Related Work}
\label{sec:related}
\paragraph{Prompt-based Learning (PL)} In general, to learn new tasks, PL prepends a task-specific template or prompt to the original input \citep{liu2021pre}. Since \citet{brown2020language} showed that a frozen GPT-3 model can achieve promising results on various in-context few-shot tasks through manually designed prompts, many efforts have been made in PL. While early efforts mainly focused on designing \emph{discrete} prompts \citep{schick-schutze-2021-exploiting,gao2020making}, more recent works attempt to learn trainable parameters, \emph{soft} prompts \citep{li2021prefix,liu2021gpt,lester2021power,qin2022lfpt}, showing impressive performance on a variety of tasks, especially with low supervision.

\paragraph{Few-shot Summarization} 
Few-shot learning remains under-explored in abstractive summarization. PEGASUS \citep{zhang2020pegasus}, which pre-trains the model on HugeNews with the \emph{gap-sentence generation objective} to predict salient sentences from the rest of the document, achieves very strong 100-shot performance on several datasets, including XSum. WikiTransfer \citep{fabbri2020improving} proposes to construct relevant pseudo-samples from Wikipedia to fine-tune pre-trained models before few-shot fine-tuning. This approach achieves great progress in zero-shot and few-shot summarization, but at the cost of using external data and performing an intermediate fine-tuning phase. \citet{bi2021boosting} add auxiliary tasks such as object prediction and entailment to the cross-entropy training objective to boost few-shot summarization performance. PSP \citep{liu2022psp} pre-trains soft prompts and then fine-tunes them for summarization, outperforming the base BART \citep{lewis2019bart} on few-shot benchmarks by a good margin. \citet{ravaut2022towards} train a second-stage BART fusion-in-decoder to combine multiple summary candidates into a new one, which helps boost the base PEGASUS performance in few-shot.

\paragraph{Guidance and Planning for Summarization}
CTRLSum \citep{he2020ctrlsum} trains the summarization model with input keywords which allow users to input any desired keywords at inference, enabling better controllability on the summarization process. GSum \citep{dou2020gsum} uses a second encoder to leverage external signals, such as keywords or salient sentences predicted by an extractive summarization model like MatchSum \citep{zhong2020extractive}, to guide the base encoder-decoder abstractive summarization model, achieving state-of-the-art results. FROST \citep{narayan2021planning} modifies the PEGASUS \citep{zhang2020pegasus} summarization pre-training objective to condition on entity chains. However, they condition through the \emph{decoder} by making the model generate first the entity chain, then the summary; while we input the entity chain to the \emph{encoder}. Besides, we use a mix of soft and discrete prompts for summary generation. To the best of our knowledge, there is no work proposing a controllable summarization model while being parameter-efficient and/or data-efficient: our model combines all three properties.

\section{Method}
\label{sec:model}
\subsection{Problem Formulation}


\begin{figure*}
    \centering 
    \includegraphics[width=0.96\textwidth]{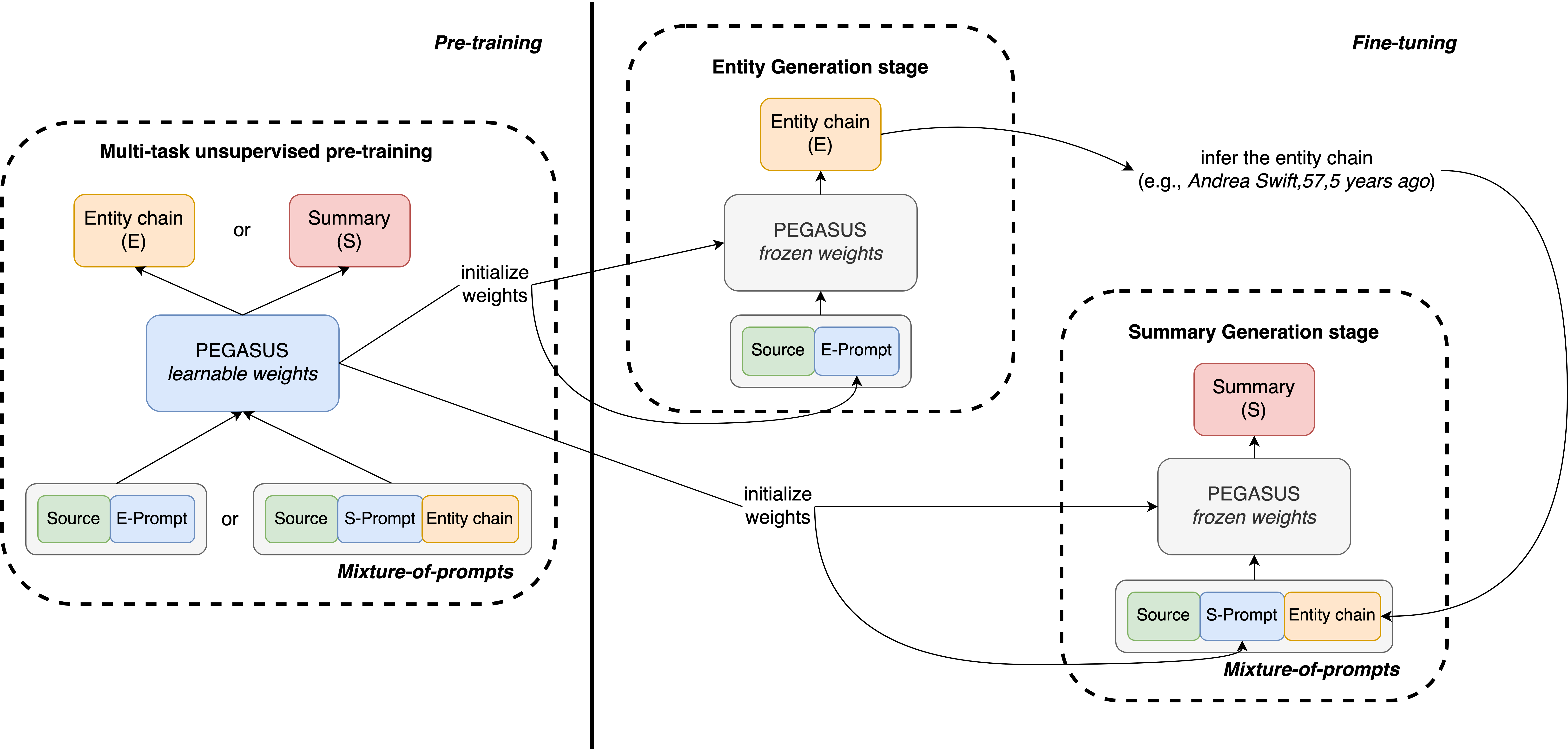}
    \caption{\small PromptSum training. In the \textbf{Pre-training} stage (left), the model alternates between the entity-chain prediction and summary prediction tasks, both with pseudo-labels and each with a dedicated input soft prompt (\emph{E-Prompt} for entity and \emph{S-Prompt} for summary). All model weights are updated in pre-training (as marked by the color blue). In the \textbf{Entity Generation} stage, PEGASUS weights (initialized from pre-training) are kept frozen (gray), and \emph{E-Prompt} is further optimized for entity chain generation with the available entity-chain supervision. In the \textbf{Summary Generation} stage, PEGASUS weights are also initialized from pre-training and kept frozen, and this time \emph{S-Prompt} is tuned for summary generation with the available summary labels. Besides, for summary generation, the model also conditions on predicted entity chains from the first prompt tuning stage.}
    \label{fig_1}
\end{figure*}

From an input text $X$, we decompose the summary (noted $Y$) generation process into two subtasks: 

\begin{itemize}[leftmargin=*]

    \item \emph{Entity Generation}: generation of a chain of salient (discrete) entities $E$ from the input $X$.

    \item \emph{Summary Generation}: generation of the summary $Y$ conditioned on \emph{both} $X$ and $E$. 
    
\end{itemize} 

By dividing the traditional summarization process into two stages, we aim to mimic a human's thinking process of generating summaries by first planning out the content with a sequence of discrete entities before writing down the final summary \citep{cao-etal-2018-retrieve}. 
Following this more practical two-stage design, we can better control the summaries with different entity chains (\Cref{sec:4_3}), as users can input any entity at inference. 

\subsection{Multi-Task Prompt Tuning}

To learn both summarization subtasks with the least amount of parameters and simplest architecture, we employ \emph{multi-task} prompt tuning, where we use the same PLM for both tasks but with different prompts. We choose PEGASUS \citep{zhang2020pegasus} as the backbone PLM, due to its strong generalization across many summarization tasks in both few-shot and full-shot. Given a training dataset $D^{tr} = \{(X_1, Y_1), \dots, (X_n, Y_n)\}$ for a task $T$ and a PLM $\theta$, prompt tuning or PT \citep{lester2021power} prepends (or appends) the input text $X_i$ with a sequence of tunable prompt tokens $P$, while all other parameters remain frozen during training. 

For the \emph{Entity Generation} task (i.e., predicting salient entities $E_{i}$ from input $X_{i}$), we first formulate a learning dataset $D^{tr}_{E} =  \{(X_1, E_1), \dots, (X_n, E_n)\}$ where $E_{i}$ represents the entity chain tagged from the ground truth summary $Y_{i}$, following \cite{narayan-etal-2021-planning}.  Then, to prompt-tune, we use soft prompt tokens $P_{E}$, parameterized by prompt embeddings $\phi_{E}$, optimized through gradient descent with the objective:

\vspace{-1.2em}
\begin{align} 
\begin{split}
\label{eq: te_objective}
    \gL_{\phi_{E}} &= \gL(\phi_E, D^{tr}_{E}) \\
    &= -\sum_{i=1}^n \log(p(E_i | [X_i, P_{E}], \phi_{E}, \theta))
\end{split}
\end{align}
\vspace{-0.5em}

where $[ , ]$ is a concatenation operation: we concatenate the soft prompt to the \emph{right} of the source. 

For \emph{Summary Generation}, generating summaries $Y_{i}$ from input $X_{i}$ and entity chain $E_{i}$, we formulate the training dataset $D^{tr}_{S} =  \{(X_1, E'_1, Y_1), \dots, (X_n, E'_n, Y_n)\}$, where $E'_{i}$ is inferred from the prompt-tuned model in the previous stage. Then, we use both the soft prompt tokens $P_{S}$ and the discrete prompt $E'_{i}$ to generate the summary $Y_{i}$. Similar to the first prompt tuning stage, only prompt embeddings $\phi_{S}$ are optimized with gradient descent through the following objective:

\vspace{-1.0em}
\begin{align} 
\begin{split}
\label{eq: ts_objective}
    \gL_{\phi_{S}} &= \gL(\phi_S, D^{tr}_{S}) \\
    &= -\sum_{i=1}^n \log(p(Y_i | [X_i, P_{S}, E'_i], \phi_{S}, \theta))
\end{split}
\end{align}
\vspace{-0.5em}

To summarize, we train each subtask through a dedicated soft prompt, and also condition on the generated entity chain as discrete prompt for the \emph{Summary Generation} subtask.

\subsection{Model Training}
\label{sec:3_3}

\paragraph{Pre-training}
To better help the model learn each task only with the corresponding soft prompt, we pre-train both soft prompts, in the same vein as PPT \cite{gu-etal-2022-ppt}. Like PEGASUS \citep{zhang2020pegasus}, we leverage the C4 dataset for pre-training, which is first introduced in \citet{raffel2019exploring}. We initialize our model, named \emph{PromptSum}, from the PEGASUS checkpoint pre-trained on C4 and HugeNews, and perform our multi-task prompt tuning pre-training on the \emph{realnewslike} subset of C4, containing 13M high-quality samples. We construct pre-training labels that are similar to \citet{narayan2021planning}: labels for \emph{Summary Generation} follow the \emph{gap-sentences generation} pre-training objective: they are salient sentences (defined as maximizing {\sc{Rouge}} with the rest of the document) chunked out from each document to form a pseudo-summary. Labels for \emph{Entity Generation} simply consist of the tagged entities found in each pseudo-summary. We initialize soft prompt embeddings for both tasks from PEGASUS embeddings of the most frequent tokens. During pre-training, within each batch, each pre-training document is used once for entity chain generation, and also once for summary generation, each time with the corresponding soft prompt being inputted to the model. The final pre-training loss is simply the sum of each subtask loss: $\gL_{\phi} = \gL_{\phi_{E}} + \gL_{\phi_{S}}$

We do not freeze the backbone model weights during pre-training: all PEGASUS weights, as well as each soft prompt weights, are updated. Such weights updating is only done once, protecting our parameter-efficiency constraint during fine-tuning.

\begin{table*}
\resizebox{\textwidth}{!}{
\begin{tabular}{lcccccccccccc}
\toprule
\multirow{2}{*}{\textbf{Dataset}} 
& \multicolumn{3}{c}{\textbf{\# Data points}} 
& \multicolumn{2}{c}{\textbf{\# Words}} 
& \multicolumn{2}{c}{\textbf{\# Tokens (PEGASUS)}} 
& \multicolumn{2}{c}{\textbf{\# Entities}} 
& \multicolumn{3}{c}{\textbf{New summary n-grams (\%)}}  \\
\cmidrule(lr){2-4}   
\cmidrule(lr){5-6} 
\cmidrule(lr){7-8}
\cmidrule(lr){9-10}  
\cmidrule(lr){11-13}  
& Train & Val & Test 
& Doc. & Summ. 
& Doc. & Summ. 
& Doc. & Summ. 
& 1-grams & 2-grams & 3-grams \\
\midrule
CNN/DM    & 287,113 & 13,334 & 11,490 & 786.68 & 55.06 & 851.53 & 64.57 & 64.13 & 5.71 & 12.07 & 51.05 & 71.38 \\
XSum      & 204,045 & 11,332 & 11,334 & 414.51 & 22.96 & 456.96 & 26.01 & 38.26 & 2.85 & 33.98 & 83.33 & 95.52 \\
BillSum   & 17,055 (ours) & 1,894 (ours) & 3,269 & 1,659.13 & 203.88 & 1,759.92 & 209.19 & 109.05 & 14.01 & 10.22 & 37.77 & 54.87 \\
SAMSum    & 14,732 & 818 & 819 & 124.07 & 23.42 & 133.07 & 25.66 & 11.45 & 3.25 & 33.88 & 79.02 & 90.10 \\
\bottomrule
\end{tabular}
}
\caption{\small Statistics on the datasets that we used for experiments. \emph{Doc.} is short for document, and \emph{Summ.} for summary. Tokens counts are calculated based on PEGASUS tokenization.}
\label{tab:1}
\end{table*}

\paragraph{Fine-tuning}
Fine-tuning the model on labeled data follows a two-stage approach. First, we extract entity chains from the available ground truth summaries and train the soft prompt for \emph{Entity Generation}. Then, we infer the entity chain, and train the soft prompt for \emph{Summary Generation}. For each tuning stage, the model is intialized with weights from pre-training, and the corresponding soft prompt is also initialized from its pre-training weights. Besides, during both pre-training and \emph{Summary Generation}, we find it beneficial to use \emph{teacher forcing for entity chains}: at training time, we input the ground truth entity chain to the model, instead of the predicted one. Not only such teacher forcing leads to higher performance, but it also speeds up training due to bypassing the need to generate entity chains on the whole training set. At inference, predicted entity chains are first inferred on the whole validation or test set. 

We refer to \Cref{fig_1} for an overview of the whole training process of PromptSum, with multi-task pre-training on the left, and the prompt-tuning stages in the center (entity generation) and right (summary generation), respectively.

\section{Experiments}
\label{sec:experiments}
\subsection{Experimental Setup}
\label{sec:4_1}

We initialize PromptSum before pre-training with the \emph{google/pegasus-large} checkpoint downloaded from HuggingFace \emph{transformers} library \citep{wolf-etal-2020-transformers}. To detect entities as ground truth, the \emph{spacy} library is used. All fine-tuning datasets used in this paper are downloaded with the HuggingFace \emph{datasets} library \citep{lhoest-etal-2021-datasets}.

Following PEGASUS \citep{zhang2020pegasus}, we use Adafactor \citep{shazeer2018adafactor} as the optimizer for all experiments. We pre-train once for 400k steps, using an effective batch size of 256 and data parallelism over 8 GPUs, evaluating every 5k optimization steps. We will release our pre-trained checkpoint publicly, from which all fine-tuning experiments were initialized. We use 100 tokens per soft prompt, each with embedding size 1024 (as per PEGASUS-large), leading to a total of 204,800 trainable parameters (sum over both soft prompts) during fine-tuning, or less than 0.1\% PEGASUS-large's total number of parameters. We cap entity chains at 100 tokens as well.

We fine-tune PromptSum on four popular summarization benchmarks

\begin{itemize}[leftmargin=*]

\item \textbf{CNN-DailyMail} \citep{hermann2015teaching}: the task is to summarize 93k news articles from CNN and 220k from DailyMail newspapers into highlighted bullet points, which form abstractive summaries with weak abstractiveness. We follow the non anonymized version from \citet{see2017get}.

\vspace{-0.5em}
\item \textbf{XSum} \citep{narayan2018don}: the task is to compress 227k BBC articles into highly abstractive, single-sentence summaries. 

\vspace{-0.5em}
\item \textbf{BillSum} \citep{kornilova2019billsum}: the task is to summarize 22k US Congressional bills from several sessions of the American Congress.

\vspace{-0.5em}
\item \textbf{SAMSum} \citep{gliwa2019samsum}: the task is to recapitulate 16k daily-life conversations. SAMSum's input dialogues are significantly shorter than the source document in other datasets.

\end{itemize}

We refer the reader to \Cref{tab:1} for general statistics on each dataset. Datasets are selected to cover multiple domains (news, legal, dialogue), compression ratios (from 5.69\% on XSum to 19.28\% on SAMSum), summary lengths (from 26 tokens on XSum and SAMSum to 209 on BillSum), and abstractiveness levels (evaluated with the fraction of new n-grams in the summaries).

We fine-tune PromptSum in both 100-shot and full-shot on each dataset. In 100-shot, we subsample three random pairs of training and validation sets each of size 100 from the training set (following the setup of \cite{perez2021true}), and fine-tune a model on each pair, then report results on the test set, averaging over the three models. We fine-tune each task for 60 epochs. In full-shot, we fine-tune entity generation for 5 epochs, and summary generation for 10 epochs on CNN/DM and XSum, 20 epochs on BillSum, and 30 on SAMSum. All hyper-parameters can be found in \Cref{sec:appendix_a}. We use beam search with a beam width of 4 for both entity chain generation and summary generation. 



\subsection{Parameter-Efficient Summarization}
\label{sec:4_2}

We evaluate PromptSum summarization performance using the standard {\sc{Rouge}} metric \citep{lin2004rouge}, in its three commonly used versions of {\sc{Rouge-1}}, {\sc{Rouge-2}}, and {\sc{Rouge-L}} (using summary-level {\sc{Rouge-L}}); as well as with BERTScore \cite{zhang2019bertscore}.

\begin{table*}[]
\resizebox{0.99\textwidth}{!}{
\begin{tabular}{llcccccccccccccccc}

\toprule 

\textbf{}                 
& \textbf{}                                           
& \multicolumn{4}{c}{\textbf{CNN/DM}}      
& \multicolumn{4}{c}{\textbf{XSum}}                        
& \multicolumn{4}{c}{\textbf{BillSum}}                     
& \multicolumn{4}{c}{\textbf{SAMSum}} \\
\cmidrule(lr){3-6} 
\cmidrule(lr){7-10} 
\cmidrule(lr){11-14} 
\cmidrule(lr){15-18} 
\textbf{Fine-tuning Size}             
& \textbf{Model}                                      
& \textbf{R-1} 
& \textbf{R-2} 
& \textbf{R-L} 
& \textbf{BS} 
& \textbf{R-1} 
& \textbf{R-2} 
& \textbf{R-L} 
& \textbf{BS} 
& \textbf{R-1} 
& \textbf{R-2} 
& \textbf{R-L} 
& \textbf{BS} 
& \textbf{R-1} 
& \textbf{R-2} 
& \textbf{R-L} 
& \textbf{BS} \\

\midrule 

\multirow{6}{*}{\textbf{100-shot}} 
& \textbf{PromptSum}*         & \cellcolor{blue!15}\textbf{37.33}$^{\dagger}$ & \cellcolor{blue!15}\textbf{15.75}$^{\dagger}$ & \cellcolor{blue!15}\textbf{33.98}$^{\dagger}$ & \cellcolor{blue!15}\textbf{87.83}$^{\dagger}$ & \cellcolor{blue!15}\textbf{41.54}$^{\dagger}$ & \cellcolor{blue!15}\textbf{18.94}$^{\dagger}$ & \cellcolor{blue!15}\textbf{33.78}$^{\dagger}$ & \cellcolor{blue!15}\textbf{91.56}$^{\dagger}$ & \cellcolor{blue!15}\textbf{37.11}$^{\dagger}$ & \cellcolor{blue!15}\textbf{16.69}$^{\dagger}$ & \cellcolor{blue!15}\textbf{27.66}$^{\dagger}$ & \cellcolor{blue!15}\textbf{84.04}$^{\dagger}$ & \cellcolor{blue!15}\textbf{41.18}$^{\dagger}$ & \cellcolor{blue!15}\textbf{17.72}$^{\dagger}$ & \cellcolor{blue!15}\textbf{33.82}$^{\dagger}$ & \cellcolor{blue!15}\textbf{91.08} \\
& w/o pre-training            & 27.70 & 10.43 & 24.63 & 80.82 & 28.99 & 10.44 & 22.50 & 88.27 & 27.87 & 10.92 & 20.80 & 81.90 & 29.32 & 10.55 & 24.78 & 87.58 \\
& w/o fine-tuning E-prompt    & 30.84 & 10.57 & 27.81 & 86.50 & 40.50 & 17.99 & 32.29 & 91.09 & 32.16 & 12.15 & 22.99 & 81.97 & 37.96 & 15.48 & 31.36 & 90.90 \\
& w/o entity chain            & 27.40 & 8.83 & 24.72 & 85.94 & 38.86 & 16.81 & 31.53 & 90.89 & 30.72 & 11.25 & 22.18 & 81.57 & 34.96 & 14.84 & 30.00 & 90.74 \\
& w/o fine-tuning S-prompt    & 29.22 & 10.68 & 25.62 & 85.13 & 27.38 & 10.67 & 21.40 & 88.11 & 32.25 & 12.86 & 23.45 & 81.77 & 20.16 & 5.09 & 17.67 & 85.19 \\
& w/o S-prompt                & 26.00 & 8.88 & 22.67 & 84.18 & 26.67 & 10.09 & 20.99 & 87.70 & 29.93 & 10.39 & 20.56 & 80.91 & 18.30 & 4.71 & 15.99 & 84.21 \\

\midrule 

\multirow{6}{*}{\textbf{Full-shot}} 
& \textbf{PromptSum}          & 40.06 & \textbf{18.34} & 36.82 & \textbf{88.59}$^{\dagger}$ & \textbf{43.31} & \textbf{20.23} & \textbf{35.36}$^{\dagger}$ & \textbf{92.00}$^{\dagger}$ & \textbf{46.70}$^{\dagger}$ & \textbf{28.69} & \textbf{35.40} & 87.96 & \textbf{46.72}$^{\dagger}$ & \textbf{22.35}$^{\dagger}$ & \textbf{38.60}$^{\dagger}$ & \textbf{91.84}$^{\dagger}$ \\
& w/o pre-training            & \textbf{40.25} & \textbf{18.43} & \textbf{37.09} & 88.41 & 43.14 & 20.12 & 34.89 & 91.74 & 44.44 & 25.32 & 33.65 & 87.23 & 33.51 & 11.47 & 27.32 & 90.06 \\
& w/o fine-tuning E-prompt    & 33.76 & 13.05 & 30.72 & 87.30 & 40.43 & 17.76 & 31.98 & 91.03 & 45.10 & 27.88 & 34.79 & 87.84 & 43.08 & 19.50 & 35.64 & 91.50 \\
& w/o entity chain            & 34.12 & 13.74 & 31.22 & 87.51 & 41.36 & 18.67 & 33.85 & 91.63 & 45.08 & \textbf{28.66} & \textbf{35.43} & \textbf{88.17} & 41.82 & 20.16 & 36.08 & 91.49 \\
& w/o fine-tuning S-prompt    & 31.75 & 12.59 & 28.01 & 85.79 & 28.04 & 11.09 & 21.88 & 88.31 & 32.40 & 13.20 & 23.71 & 81.84 & 20.82 & 5.34 & 18.03 & 85.39 \\
& w/o S-prompt                & 27.71 & 10.00 & 24.22 & 84.62 & 26.93 & 10.22 & 21.28 & 87.78 & 29.90 & 10.45 & 20.59 & 80.92 & 19.42 & 4.96 & 16.50 & 84.15 \\

\bottomrule 
      
\end{tabular}
}
\caption{\small Parameter-efficient controllable summarization results. *Blue cells correspond to our PromptSum proposed model which is of low supervision, parameter-efficient and controllable. Rows below \textbf{PromptSum} correspond to an ablation of the model's components. $^{\dagger}$marks indicate that PromptSum results are significantly better ($p$-value of paired t-test smaller than 0.05) than \emph{all} the other baselines. \textbf{R-1/2/-L} stands for ROUGE-1/2/L, \textbf{BS} is the BERTScore. Best numbers within 0.1 are in bold.}
\label{tab:2}
\end{table*}

We report 100-shot and full-shot summarization results in \Cref{tab:2}. To validate every component of PromptSum's pipeline displayed in \Cref{fig_1}, we include a thorough ablation study with several model variants:

\begin{itemize}[leftmargin=*]

\vspace{-0.5em}
\item \emph{w/o pre-training}: we by-pass the pre-training stage and directly use PEGASUS-large as a backbone, and fine-tune both E-prompt and S-prompt.

\vspace{-0.5em}
\item \emph{w/o fine-tuning E-prompt}: at inference, we use the E-prompt from the pre-training stage (not fine-tuned) to generate the entity chain, and keep the fine-tuned S-prompt. 

\vspace{-0.5em}
\item \emph{w/o entity chain}: at inference, we remove the entity chain, but keep the fine-tuned S-prompt. 

\vspace{-0.5em}
\item \emph{w/o fine-tuning S-prompt}: at inference, we use the S-prompt from the pre-training stage (not fine-tuned), but we keep entity chains obtained with the fine-tuned E-prompt. 

\vspace{-0.5em}
\item \emph{w/o S-prompt}: at inference, we remove S-prompt entirely, but keep the entity chain. 

\end{itemize}


\begin{table}[]
\resizebox{\columnwidth}{!}{
\begin{tabular}{llccccc}

\toprule 

\textbf{Fine-tuning Size}     
& \textbf{Model}      
& \textbf{Control.?} 
& \textbf{\begin{tabular}[c]{@{}c@{}}Param.-\\ efficient?\end{tabular}} 
& \textbf{R-1} 
& \textbf{R-2} 
& \textbf{R-L} \\

\midrule 

\multirow{5}{*}{\textbf{100-shot}} 

& PEGASUS-large \citep{zhang2020pegasus}          & \xmark & \xmark & 39.07 & 16.44 & 31.27 \\
& WikiTransfer \citep{fabbri-etal-2021-improving} & \xmark & \xmark & 37.26 & 14.20 & 28.85 \\
& SummaFusion \citep{ravaut2022towards}           & \xmark & \xmark & 39.86 & 17.01 & 31.68 \\

\cdashline{2-7}

& PSP* \citep{liu2022psp}                          & \xmark & \checkmark & 32.50 & 10.83 & 25.03 \\

\cdashline{2-7}

& \textbf{PromptSum}                               & \checkmark & \checkmark & \textbf{41.54} & \textbf{18.94} & \textbf{33.78}  \\

\midrule

\multirow{7}{*}{\textbf{Full-shot}} 

& T5-large** \citep{raffel2019exploring}           & \xmark & \xmark & 44.28 & 21.01 & 36.14 \\
& BART-large \citep{lewis2019bart}                & \xmark & \xmark & 45.14 & 22.27 & 37.25 \\
& PEGASUS-large \citep{zhang2020pegasus}          & \xmark & \xmark & 47.21 & 24.56 & 39.25 \\

\cdashline{2-7}

& FROST \citep{narayan-etal-2021-planning}        & \checkmark & \xmark & \textbf{47.80} & \textbf{25.06} & \textbf{39.76} \\

\cdashline{2-7}

& Prefix-Tuning 2\% \citep{li2021prefix}          & \xmark & \checkmark & 43.80 & 20.93 & 36.05 \\
& Prefix-Tuning 0.1\% \citep{li2021prefix}        & \xmark & \checkmark & 42.92 & 20.03 & 35.05 \\

\cdashline{2-7}

& \textbf{PromptSum}                              & \checkmark & \checkmark & 43.34 & 20.32 & 35.41 \\

\bottomrule
        
\end{tabular}
}
\caption{Abstractive summarization {\sc{Rouge}} results on XSum. \textbf{Control.?} indicates whether the model is controllable, and \textbf{Param.-efficient} whether the model only fine-tunes a small subset of parameters. *Exact numbers were obtained by emailing the authors. **Results correspond to our re-implementation. \textbf{R-1/2/-L} stands for ROUGE-1/2/L. Best numbers within 0.1 are in bold.}
\label{tab:3}
\end{table}

As seen in \Cref{tab:2}, all model components from \Cref{fig_1}, namely pre-training, fine-tuning both soft prompts, and conditioning on the generated entity chain and the S-prompt, are compulsory for PromptSum to reach full abstractive summarization peformance. In particular, without pre-training, PromptSum collapses in few-shot (-10.82 {\sc{Rouge-1}} on average), as well as on SAMSum in full-shot (-13.21 {\sc{Rouge-1}}). Removing the entity chain harms {\sc{Rouge-1}} by -6.31 points on average. 

\begin{table}[]
\resizebox{\columnwidth}{!}{
\begin{tabular}{llcccccc}

\toprule 

\textbf{Dataset}                  
& \textbf{Fine-tuning Size} 
& \textbf{R-1} 
& \textbf{R-2} 
& \textbf{R-L} 
& \textbf{Prec.}
& \textbf{Rec.}
& \textbf{F-1} \\

\midrule 

\multirow{3}{*}{\textbf{CNN/DM}}  
& 0-shot (just pre-training) & 21.67 & 7.61 & 20.52 & 32.64 & 18.75 & 21.36 \\
& 100-shot                   & 41.19 & 22.05 & 36.18 & 41.34 & 34.06 & 34.91 \\
& Full-shot                  & \textbf{47.01} & \textbf{25.92} & \textbf{41.20} & \textbf{45.86} & \textbf{39.79} & \textbf{40.46} \\

\midrule 

\multirow{3}{*}{\textbf{XSum}}    
& 0-shot (just pre-training) & 24.42 & 11.43 & 23.20 & 26.41 & \textbf{37.62} & 28.51 \\
& 100-shot                   & 41.03 & 20.22 & 39.03 & \textbf{44.61} & 28.86 & 33.09 \\
& Full-shot                  & \textbf{44.78} & \textbf{22.36} & \textbf{42.06} & 43.86 & 33.12 & \textbf{36.10} \\

\midrule 

\multirow{3}{*}{\textbf{BillSum}} 
& 0-shot (just pre-training) & 16.16 & 7.22 & 15.85 & 31.14 & 6.37 & 9.24 \\
& 100-shot                   & 37.22 & 23.09 & 33.45 & 37.76 & 24.99 & 26.37 \\
& Full-shot                  & \textbf{42.25} & \textbf{28.74} & \textbf{38.10} & \textbf{40.89} & \textbf{28.81} & \textbf{30.03} \\

\midrule 

\multirow{3}{*}{\textbf{SAMSum}}  
& 0-shot (just pre-training) & 29.72 & 9.30 & 28.77 & \textbf{67.11} & 46.38 & 51.14 \\
& 100-shot                   & 54.38 & 18.05 & 49.33 & 64.29 & 58.01 & 58.05 \\
& Full-shot                  & \textbf{57.80} & \textbf{21.32} & \textbf{52.64} & 65.03 & \textbf{62.13} & \textbf{60.73} \\

\bottomrule

\end{tabular}
}
\caption{\small Evaluation of the entity chain generation with the PromptSum's E-prompt. \textbf{R-1/2/-L} stands for ROUGE-1/2/L, \textbf{Prec.} and \textbf{Rec.} are precision and recall, respectively. Best numbers within 0.1 are in bold.}
\label{tab:4}
\end{table}

To get a better picture of PromptSum performance, we include a comparison with other abstractive summarization results on XSum (arguably the most popular benchmark in abstractive summarization) in \Cref{tab:3}. We split models in categories following whether the model is controllable, parameter-efficient, or none, with PromptSum being the only method endowed with both properties. PromptSum outperforms parameter-efficient methods Prefix-Tuning 0.1\% in full-shot and PSP in 100-shot, despite also being controllable. In 100-shot, it is even better than full-parameters SOTA models PEGASUS and SummaFusion. In full-shot, PromptSum underperforms full-parameter methods, which echoes other work reporting soft prompt tuning results for abstractive summarization \citep{li2021prefix, ding2022delta}, yet it stays within 10\% of SOTA method FROST, but at several orders of magnitude less trainable parameters budget.

We then evaluate the performance of the entity generation task in \Cref{tab:4}, comparing the generated and ground truth entity chains both with the {\sc{Rouge}} metrics, and with precision, recall and F-1. The latter three metrics provide a complementary view to {\sc{Rouge}} since they do not penalize entities predicted in the wrong order. PromptSum is able to achieve strong scores ({\sc{Rouge-1}} above 40.0) on all setups even with just 100-shot supervision. \Cref{tab:2} ablations \emph{w/o finetuning E-prompt} and \emph{w/o entity chain} showed that not using a good entiy chain is damaging to summary generation performance. Overall, the entity chain quality drives the performance of the final summary, a relationship between both tasks which we further illustrate in \Cref{sec:appendix_b}.

\begin{table}[t]
\resizebox{\columnwidth}{!}{
\begin{tabular}{llcccccccc}
\toprule

\textbf{Dataset}        
& \textbf{Model}              
& \textbf{K=1} 
& \textbf{K=2} 
& \textbf{K=5} 
& \textbf{Oracle ents.} \\

\midrule

\multirow{3}{*}{\textbf{CNN/DM}} 
& CTRLSum                     & 52.0 & 33.2 & 5.9 & 25.3 \\
& PromptSum - no pre-training & 25.8 & 9.2 & 3.4 & 6.1 \\
& PromptSum                   & \textbf{89.1} & \textbf{75.6} & \textbf{41.3} & \textbf{60.6} \\

\midrule

\multirow{3}{*}{\textbf{XSum}} 
& CTRLSum                     & 21.0 & 5.9 & 0.0 & 23.7 \\
& PromptSum - no pre-training & 23.4 & 17.9 & 12.0 & 51.6 \\
& PromptSum                   & \textbf{77.6} & \textbf{56.4} & \textbf{17.2} & \textbf{89.3} \\

\midrule

\multirow{3}{*}{\textbf{BillSum}} 
& CTRLSum                     & 35.6 & 13.7 & 0.5 & 11.1 \\
& PromptSum - no pre-training & 38.3 & 18.8 & 4.9 & 5.0 \\
& PromptSum                   & \textbf{53.2} & \textbf{35.5} & \textbf{20.0} & \textbf{9.0} \\

\midrule

\multirow{3}{*}{\textbf{SAMSum}} 
& CTRLSum                     & 60.3 & 36.0 & 1.3 & 57.7 \\
& PromptSum - no pre-training & 47.7 & 35.0 & 11.9 & 59.8 \\
& PromptSum                   & \textbf{75.1} & \textbf{64.3} & \textbf{41.9} & \textbf{86.6} \\

\bottomrule

\end{tabular}
}
\caption{\small Controllabilty results (Success rate \%) across all four datasets, in 100-shot. A controllability experiment is \emph{successful} if the summary mentions all entities in the given entity-chain. \textbf{K=n} denotes entity chain including $n$ unique entities randomly selected from source input. \textbf{Oracle entities} is the entity chain extracted from the ground truth. Best numbers within 0.1 are in bold.}
\label{tab:5}
\end{table}

\subsection{Controllability} 
\label{sec:4_3}

To test controllability on entities, we perform the following experiments with intervention on the entity chain: we sample 100 documents from the test set, and randomly choose $K$ entities from the source or use oracle entities extracted from ground truth summaries, then compute the success rate that all the selected entities appear in the generated summaries. We report controllability results of PromptSum and its variant without pre-training in \Cref{tab:5}. We compare it with CTRLsum \citep{he2020ctrlsum}, which also conditions on entities at the input of the encoder. We fine-tune CTRLSum ourselves using the public pre-trained checkpoint with \emph{fairseq} package.

We can see that PromptSum is significantly more controllable than CTRLsum in all settings. When $K$ increases, CTRLsum quickly loses its ability to condition on all entities while PromptSum is able to maintain a much stronger performance. This shows that PromptSum is significantly better at conditioning on multiple entities than CTRLsum while having three orders of magnitude less trainable parameters.
Besides, without pre-training, PromptSum generally becomes less effective in conditioning on entities, reaffirming our conclusion from \Cref{sec:4_2} that pre-training is critical for PromptSum.

\begin{table}[t]
\resizebox{\columnwidth}{!}{
    \begin{tabular}{lcccccc}
        \toprule
        \textbf{Dataset} 
        & \textbf{Entities} 
        & \textbf{Data\%} 
        & \textbf{Controlled} 
        & \textbf{Hal. \%} ($\downarrow$)
        & \textbf{Mean R} ($\uparrow$) 
        & \textbf{F-1} ($\uparrow$) \\
        
        \midrule
        
        \multirow{3}{*}{\textbf{CNN/DM}} &
        \emph{Non-Hal.} & 55.1 & \emph{N}  &  \emph{1.4} & \emph{29.6} & \emph{36.6} \\
        \cdashline{2-6}
        & \multirow{2}{*}{Hal.} & \multirow{2}{*}{44.9} & N  & 11.2 & \textbf{29.8} & \textbf{32.8} \\
        &  & & Y &  \textbf{2.7} & 28.7 & 32.5 \\
        
        \midrule
        
        \multirow{3}{*}{\textbf{XSum}} 
        & \emph{Non-Hal.} & 36.3 & \emph{N}  & \emph{8.9} & \emph{29.1} & \emph{38.7} \\
        \cdashline{2-6}
        & \multirow{2}{*}{Hal.} & \multirow{2}{*}{63.7} & N & 73.7 & \textbf{32.6} & \textbf{35.9}   \\
        & & & Y  & \textbf{59.6} & 31.5 & 32.5   \\
        
        \midrule
        
        \multirow{3}{*}{\textbf{BillSum}} 
        & \emph{Non-Hal.} & 3.3 & \emph{N}  & \emph{30.4} & \emph{29.6} & \emph{28.6}  \\
        \cdashline{2-6}
        & \multirow{2}{*}{Hal.} & \multirow{2}{*}{96.7} & N  & 35.1 & \textbf{30.7} & \textbf{27.1}   \\
        & & & Y  & \textbf{33.8} & 29.6 & 23.22   \\
        
        \midrule
        
        \multirow{3}{*}{\textbf{SAMSum}} 
        & \emph{Non-Hal.} & 32.7 & \emph{N} & 2.9 & 28.5 & 62.4 \\
        \cdashline{2-6}
        & \multirow{2}{*}{Hal.} & \multirow{2}{*}{67.2} & N  & 22.2  & \textbf{29.7} & \textbf{55.9}  \\
        & & & Y  & \textbf{14.0} & 27.7 & 47.0  \\
        
        \bottomrule
    \end{tabular}
    }
    \vspace{-0.5em}
    \caption{\small Hallucination results on each dataset with PromptSum trained in 100-shot. \textbf{Data\%} indicates the percentage of data points in each subset. \textbf{Hal. \%} stands for Hallucination, and lower is better. \textbf{F-1} is computed between generated entities and entities in ground truth summaries. Best numbers within 0.1 are in bold.
    \label{tab:6}}
\end{table}

To better understand how PromptSum conditions on multiple entities, we show some qualitative example in Table \ref{tab:7} on XSum (and \Cref{sec:appendix_e} on the other datasets), made by prompting with entity chains sampled from the source entities. The examples demonstrate that PromptSum can produce summaries focusing on different aspects, conditioned on the given entity chain, while still producing fluent, factual and informative summaries. This controllability trait has the potential of being useful in many downstream tasks, such as producing summaries on different topics, expanding summary candidates, and controlling hallucinations. We will next investigate controlling hallucinations as a critical use case for abstractive summarization.

\subsection{Hallucinations Control}
\label{sec:4_4}

Hallucinations, \textit{i.e.,} generating entities that do not appear in the inputs, are becoming an important threat to factuality in abstractive summarization. Detecting \citep{maynez2020faithfulness}, preventing \citep{narayan2021planning} and analyzing \citep{cao2022hallucinated} hallucinations remain major open research problems. We now show that the controllable PromptSum model has the potential of mitigating hallucinations, even under strong parameter-efficiency and data-efficiency constraints.

\begin{table*}[]
\centering
\resizebox{\textwidth}{!}{
\begin{tabular}{p{4cm}p{14cm}}
\toprule
\textbf{Entity Chain} 
& \textbf{Summary} \\

\midrule

\_ & \small \textbf{Source} Teams discussed the failure of the new elimination format with F1 commercial boss Bernie Ecclestone and Jean Todt, president of governing body, the FIA. Mercedes F1 boss Toto Wolff said Todt and Ecclestone refused to revert to the 2015 system despite teams' wishes to do so. A new aggregate system proposed by the FIA will be discussed again next week. Lewis Hamilton qualified fastest for Sunday's race in Bahrain, but there were quiet periods near the end of the first and second sessions when there were no cars out on the track. Red Bull team principal Christian Horner said all parties had agreed that the new system introduced on the eve of the 2016 season - where the slowest car is eliminated every 90 seconds in three sessions - was not the way forward. Horner said: "There is an unwillingness from the promoter and FIA to go back to 2015. The teams would go back. A compromise has been put on the table now for the teams to consider. "Let's have a look at what's been out on the table today. The bottom line is if we don't agree to a compromise, then we're stuck with what we've got and everybody agrees that what we've got isn't right." Horner and Wolff both said they did not know [...] \\

\cdashline{1-2}

\_ & \small \textbf{Ground Truth Summary} Formula 1 bosses have failed to agree on a new format for qualifying after a meeting at the Bahrain Grand Prix. \\

\cdashline{1-2}

\small \textcolor{ForestGreen}{2015} 
& \small Formula One teams have called for a return to the \textcolor{ForestGreen}{2015}  qualifying system after failing to agree on a compromise. \\

\cdashline{1-2}

\small \textcolor{Bittersweet}{Jean Todt}, \textcolor{ForestGreen}{2015} 
& \small \textcolor{Bittersweet}{Jean Todt}, president of governing body FIA, refused to revert to the \textcolor{ForestGreen}{2015}  system despite teams' wishes to do so. \\

\cdashline{1-2}

\small \textcolor{Brown}{Bernie Ecclestone}, \textcolor{ForestGreen}{2015} 
& \small \textcolor{Brown}{Bernie Ecclestone} and Jean Todt refused to revert to the \textcolor{ForestGreen}{2015} elimination format despite teams' wishes to do so. \\

\cdashline{1-2}

\small \textcolor{BurntOrange}{Red Bull}, \textcolor{JungleGreen}{today} 
& \small \textcolor{BurntOrange}{Red Bull} says a compromise has been put on the table after Formula One teams failed to agree on a new qualifying system \textcolor{JungleGreen}{today} . \\

\cdashline{1-2}

\small \textcolor{ForestGreen}{2015}, \textcolor{TealBlue}{Saturday}, \textcolor{Rhodamine}{three} 
& \small Formula One teams have called for a return to the \textcolor{ForestGreen}{2015} qualifying system after \textcolor{TealBlue}{Saturday}'s \textcolor{Rhodamine}{three} practice sessions.\\

\bottomrule

\end{tabular}
}
\caption{\small Controllability qualitative examples on XSum 100-shot. Entities are sampled from source entities.}
\label{tab:7}
\end{table*}

To test this hypothesis, we conduct the following experiments: given the prompts learned from 100-shot entity and summary tuning, we divide the test set $D^{ts} = \{(X_1, E'_1, Y_1), \ldots (X_n, E'_n, Y_n)\}$ into two subsets: a set where the predicted salient entities $E'$ are not hallucinated ($D^{ts}_{nh}$), and another set where at least one entity of $E'$ is hallucinated ($D^{ts}_{h}$). Then, we perform summarization inference on both sets as-is and document the percentage of hallucinated entities in the predicted summaries. To test PromptSum's ability to \emph{reduce hallucinations}, we also infer on $D^{ts}_{h}$ with non-hallucinated entity chains, which are produced by manually removing hallucinated entities from $E'$. If the hypothesis holds, the hallucination percentage should decrease after such controlling.

The hallucination experiment results are shown in \Cref{tab:6}. $D^{ts}_{h}$, the set where the prompt-tuned entity generation system generated hallucinated entities, represents a significant fraction of each dataset (up to 96.7\% for BillSum). As expected, on all four datasets, $D^{ts}_{h}$ contains a significantly higher percentage of hallucinations in generated summaries: 11.2\% vs 1.4\% on CNN/DM. This behavior shows that the summarization model has explicitly learned to include entity prompts in generated summaries, which further supports the observations in \Cref{sec:4_3}. Then, after manually removing the hallucinated entities from entity chain prompts $E'$ (shown as Controlled $Y$ in the table), the hallucination percentage decreased significantly (by a relative drop of -75.9\% on CNN/DM and -36.9\% on SAMSum) on all four datasets, showing promising abilities to control hallucinations. Furthermore, output summaries with a controlled entity chain maintain a {\sc{Rouge}} on par with the one of summaries without hallucinations within the entity chain, showing great promise towards achieving high quality, low hallucinations summaries despite the high noise level in training datasets \citep{goyal-durrett-2021-annotating}.

\subsection{Human Evaluation} 
\label{sec:4_5}

\begin{table}[t]

\resizebox{\columnwidth}{!}{
\begin{tabular}{llcccc}

\toprule 

\textbf{Dataset}                
& \textbf{Summary Aspect}       
& \textbf{PEGASUS} 
& \textbf{PromptSum} 
& \textbf{Tie} 
& \textbf{Agreement}* \\

\midrule 

\multirow{5}{*}{\textbf{CNN/DM}} 
& Informative          & 19.50 (3.51) & \textbf{24.25} (3.40) & 6.25 (3.59)  & 65.00 \\
& Factually consistent & 11.50(5.80) & 13.25 (11.09) & \textbf{16.25} (14.84) & 63.00 \\
& Relevant             & 21.00 (5.35) & \textbf{24.25} (3.40) & 4.75 (4.11) & 67.50 \\
& Fluent               & 19.25 (5.56) & \textbf{22.00} (4.97) & 8.75 (7.27) & 66.50 \\
& Coherent             & 18.75 (7.68) & \textbf{19.00} (8.04) & 12.25 (12.34) & 59.50 \\

\midrule 

\multirow{4}{*}{\textbf{XSum}}   
& Informative          & \textbf{21.00} (4.83) & 20.50 (1.29) & 8.50 (6.03) & 62.00 \\
& Factually consistent & 15.50 (6.76) & 15.50 (5.45) & \textbf{19.00} (10.86) & 59.50 \\
& Relevant             & 17.00 (6.78) & \textbf{18.75} (4.35) & 14.25 (10.11) & 59.00 \\
& Fluent               & 15.50 (8.70) & 16.50 (4.65) & \textbf{18.00} (13.34) & 59.00 \\

\bottomrule 

\end{tabular}
}
\caption{\small Human evaluation of abstractive summarization quality on CNN/DM and XSum. All numbers are \emph{Mean (std)} across raters of counts over the 50 points. We ignore \emph{Coherence} on XSum due to the single-sentence structure of summaries. *We follow the procedure in \citep{durmus2020feqa}.}
\label{tab:8}
\end{table}

To better understand the quality of the summaries generated by PromptSum, we ask four humans to compare outputs generated by PromptSum and PEGASUS on 50 randomly sampled test data points on both CNN/DM and XSum, with both models fine-tuned on 100 data points. The evaluators are volunteer graduate students with a full professional English competency, and are shown the source text alongside both summaries. Results of this human evaluation are shown in \Cref{tab:8}. Both models are on par with regards to factual consistency and coherence (on CNN/DM), but summaries produced by PromptSum are more relevant, as well as more informative on CNN/DM. We believe that these results are due to the additional entity chain, making the generated summary more likely to cover key aspects embodied through entities.

\section{Conclusion}
\label{sec:conclusion}


We introduced PromptSum, the first model enabling controllability in abstractive summarization while being parameter-efficient and/or only using few-shot supervision. Parameter-efficiency is achieved through the use of soft prompts, and PromptSum is rendered operational thanks to the tailored multi-task pre-training objective which we introduced. 

PromptSum provides an easy-to-use and strong control mechanism of the summary through the input entity chain. Besides, pruning hallucinated entities from the input chain drastically reduces hallucinations in the output summary. The {\sc{Rouge}} performance stays competitive, especially in few-shot, where it can match the best reported approached on XSum. Humans find PromptSum summaries more relevant than PEGASUS ones. Future work includes studying the use of other signals than entities (e.g., salient sentences) in the discrete prompt.

\bibliography{anthology,custom}
\bibliographystyle{acl_natbib}

\appendix

\section{Hyper-parameters}
\label{sec:appendix_a}
In \Cref{tab:a_1} and \Cref{tab:a_2}, we list hyper-parameters used to fine-tune PromptSum on each dataset in full-shot and few-shot, respectively. In few-shot, we evaluate every epoch, hence the absence of an \textbf{Eval steps} column. All models are trained with the Adafactor optimizer, in pytorch.

\begin{table}[h!]
\resizebox{0.99\columnwidth}{!}{
\begin{tabular}{lcccccc}
\toprule 

\textbf{Dataset} 
& \textbf{\begin{tabular}[c]{@{}c@{}}Epochs\\ Entity\end{tabular}} 
& \textbf{\begin{tabular}[c]{@{}c@{}}Epochs\\ Summary\end{tabular}} 
& \textbf{\begin{tabular}[c]{@{}c@{}}Effective\\ BS\end{tabular}} 
& \textbf{\begin{tabular}[c]{@{}c@{}}LR \\ Fine-tuning\end{tabular}} 
& \textbf{\begin{tabular}[c]{@{}c@{}}LR\\ Prompt-tuning\end{tabular}} 
& \textbf{\begin{tabular}[c]{@{}c@{}}Eval\\ steps\end{tabular}} \\

\midrule 

CNN/DM           
& 5 & 10 & 256 & 5e-5 & 5e0 & 500 \\

XSum             
& 5 & 10 & 256 & 1e-4 & 5e-3 & 500 \\

BillSum          
& 5 & 20 & 256 & 2e-4 & 5e-1 & 50  \\

SAMSum           
& 5 & 30 & 256 & 1e-4 & 5e-3 & 50  \\

\bottomrule

\end{tabular}
}
\caption{\small Full-shot fine-tuning hyper-parameters on the four datasets. \textbf{Entity} designates the entity generation prompt-tuning phase, and \textbf{Summary} the summary generation prompt-tuning phase. \textbf{BS} is the batch size, \textbf{LR} the learning rate, \textbf{Eval steps} the number of optimization steps between two consecutive evaluations. We report \textbf{LR Fine-tuning} corresponding to the learning rate used when tuning all model weights. \textbf{LR Prompt-tuning} is used for both entity generation and summary generation.}
\label{tab:a_1}
\end{table}

\begin{table}[h!]
\resizebox{0.99\columnwidth}{!}{
\begin{tabular}{lcccccc}
\toprule 

\textbf{Dataset} 
& \textbf{\begin{tabular}[c]{@{}c@{}}Epochs\\ Entity\end{tabular}} 
& \textbf{\begin{tabular}[c]{@{}c@{}}Epochs\\ Summary\end{tabular}} 
& \textbf{\begin{tabular}[c]{@{}c@{}}Effective\\ BS\end{tabular}} 
& \textbf{\begin{tabular}[c]{@{}c@{}}LR \\ Fine-tuning\end{tabular}} 
& \textbf{\begin{tabular}[c]{@{}c@{}}LR\\ Prompt-tuning\end{tabular}} \\

\midrule 

CNN/DM           
& 60 & 60 & 8 & 5e-5 & 5e-3  \\

XSum             
& 60 & 60 & 8 & 5e-5 & 5e-3  \\

BillSum          
& 60 & 60 & 8 & 5e-5 & 5e-3  \\

SAMSum           
& 60 & 60 & 8 & 5e-5 & 5e-3  \\

\bottomrule

\end{tabular}
}
\caption{\small Few-shot fine-tuning hyper-parameters on the four datasets. \textbf{Entity} designates the entity generation prompt-tuning phase, and \textbf{Summary} the summary generation prompt-tuning phase. \textbf{BS} is the batch size, \textbf{LR} the learning rate. We report \textbf{LR Fine-tuning} corresponding to the learning rate used when tuning all model weights. \textbf{LR Prompt-tuning} is used for both entity generation and summary generation.}
\label{tab:a_2}
\end{table}

\begin{table}[h!]
\resizebox{0.99\columnwidth}{!}{
\begin{tabular}{lccccccc}
\toprule 

\textbf{Dataset} 
& \textbf{\begin{tabular}[c]{@{}c@{}}Max source\\ tokens\end{tabular}} 
& \textbf{\begin{tabular}[c]{@{}c@{}}Max target\\ tokens\end{tabular}} 
& \textbf{\begin{tabular}[c]{@{}c@{}}Beam\\ width\end{tabular}} 
& \textbf{\begin{tabular}[c]{@{}c@{}}Length\\ penalty\end{tabular}} 
& \textbf{\begin{tabular}[c]{@{}c@{}}Repetition\\ penalty\end{tabular}} \\

\midrule 

CNN/DM           
& 768 & 128 & 4 & 1.0 & 1.0 \\

XSum             
& 768 & 64 & 4 & 1.0 & 1.0 \\

BillSum          
& 1024 & 256 & 4 & 1.0 & 1.0 \\

SAMSum           
& 512 & 64 & 4 & 1.0 & 1.0 \\

\bottomrule

\end{tabular}
}
\caption{\small Generation hyper-parameters on the four datasets.}
\label{tab:a_3}
\end{table}

We show hyper-parameters used to generate summaries in \Cref{tab:a_3}. All generation is done through beam search decoding. For T5 fine-tuning reported in \Cref{tab:3}, we follow the same parameters, with the exception of using 2.5 for repetition penalty.

\section{Relationship between entity and summary generation}
\label{sec:appendix_b}
\begin{table}[h]
\resizebox{\columnwidth}{!}{
\begin{tabular}{lcccc}

\toprule

\textbf{Dataset}   & CNN/DM & XSum  & BillSum & SAMSum \\

\midrule 

\textbf{100-shot}  & 0.450  & 0.381 & 0.279   & 0.294  \\
\textbf{Full-shot} & 0.488  & 0.383 & 0.261   & 0.279 \\

\bottomrule

\end{tabular}
}
\caption{\small Correlation between entity {\sc{Rouge}} and summary {\sc{Rouge}}. We compute the Pearson correlation between the mean {\sc{Rouge}} of each task across the test set of each dataset.}
\label{tab:b}
\end{table}


\begin{figure}[h]
    \centering 
    \includegraphics[width=\columnwidth]{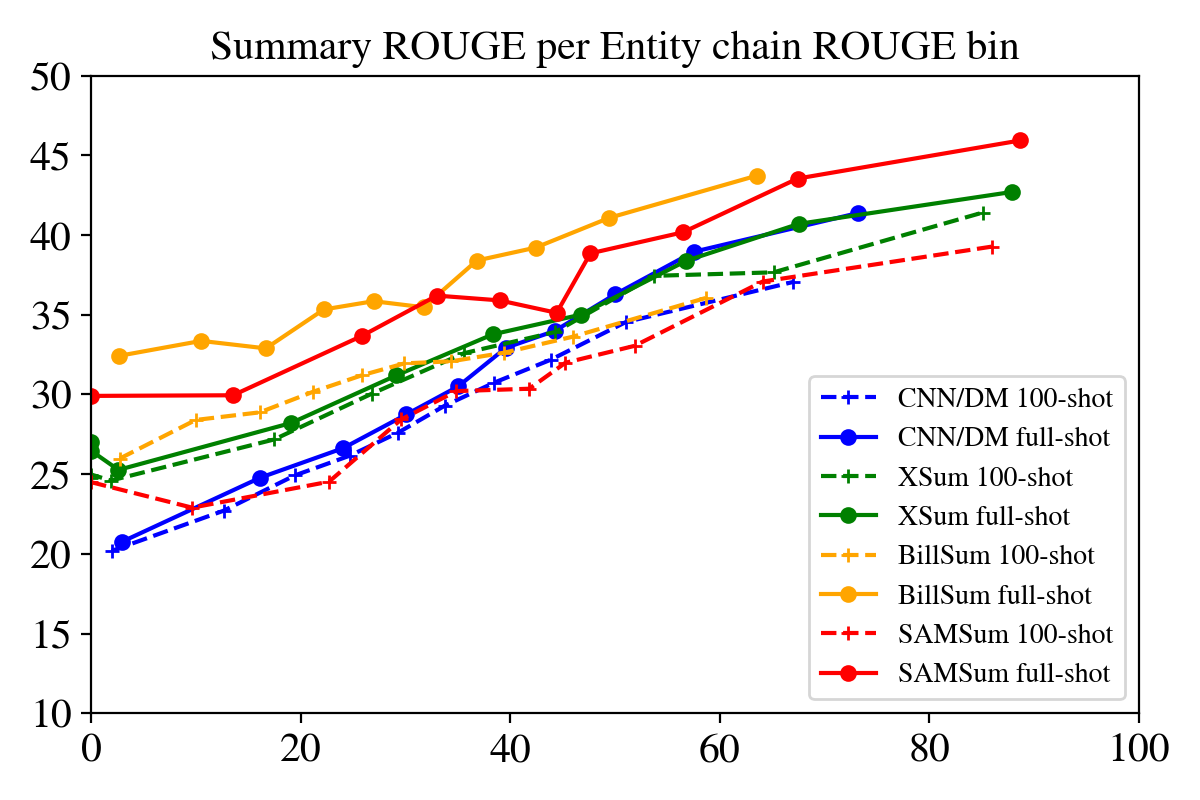}
    \caption{\small Summary {\sc{Rouge}} plotted against the entity {\sc{Rouge}}. The x-axis corresponds to the mean entity {\sc{Rouge}} of 10 equal-size bins, and the y-axis to the corresponding mean summary {\sc{Rouge}} within each entity {\sc{Rouge}} bin.}
    \label{fig_d}
\end{figure}

To cast light on how the entity chain affects the summary generation performance, in \Cref{fig_d}, we plot the mean {\sc{Rouge}} of summary generation against the mean {\sc{Rouge}} of entity generation. Specifically, we sort the test set of each dataset per increasing entity {\sc{Rouge}}, then split data points into 10 equal-size bins. The mean entity {\sc{Rouge}} in each bin forms the x-axis coordinate, and the mean Summary {\sc{Rouge}} in the corresponding bin is the y-axis coordinate. We see strong positive correlation between each {\sc{Rouge}} value: \textbf{when the entity {\sc{Rouge}} increases, so does the summary {\sc{Rouge}} on average, showing that summary quality is intertwined with entity chains quality}. 

\Cref{tab:b} confirms a high correlation (as measured per the Pearson correlation coefficient) between both tasks on all datasets. Interestingly, the correlation is not weaker in 100-shot (apart on CNN/DM), despite the model learning each task with much less information.

\section{Abstractiveness}
\label{sec:appendix_c}

\begin{figure}[h]
    \centering 
    \includegraphics[width=\columnwidth]{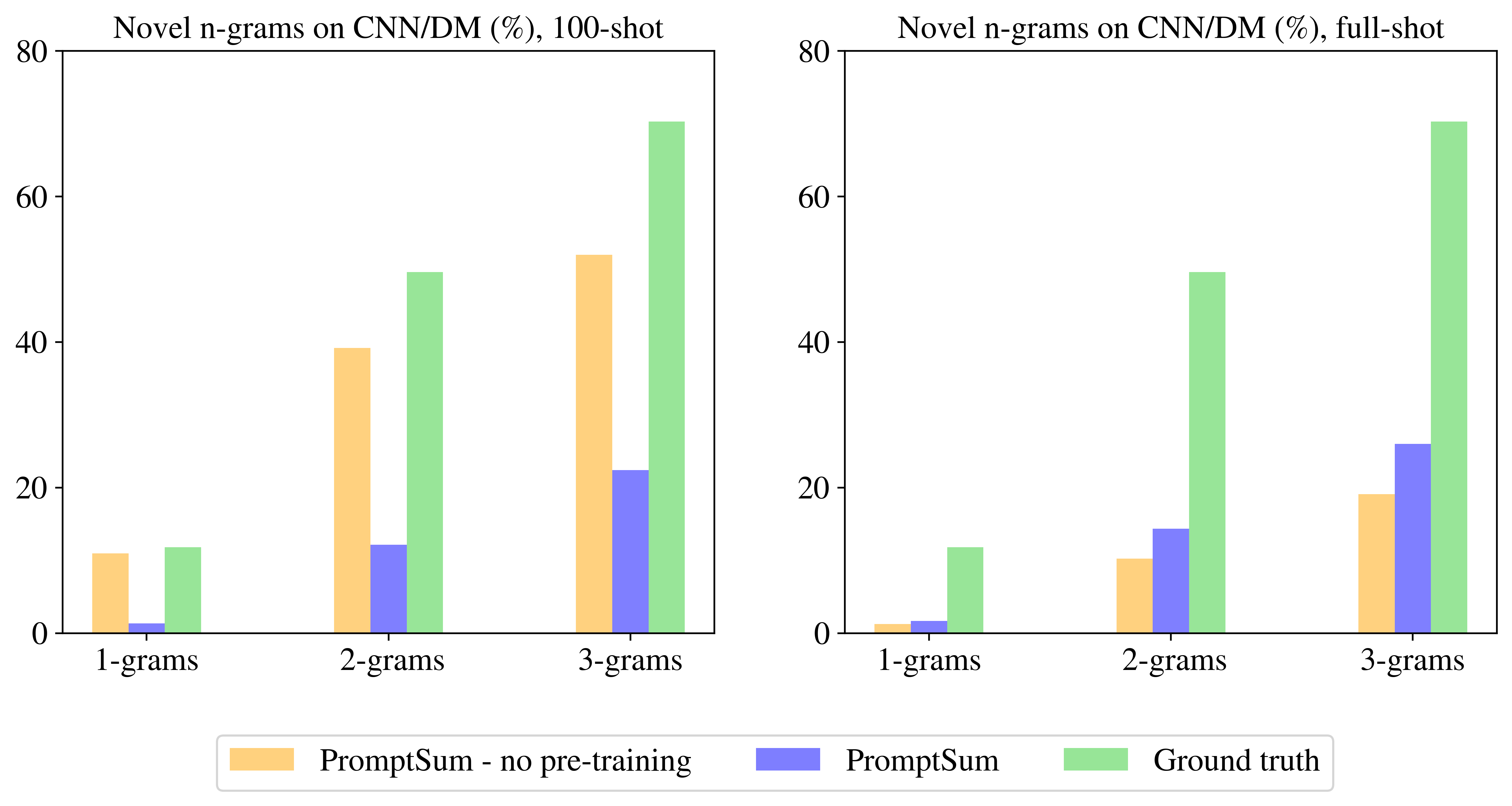}
    \includegraphics[width=\columnwidth]{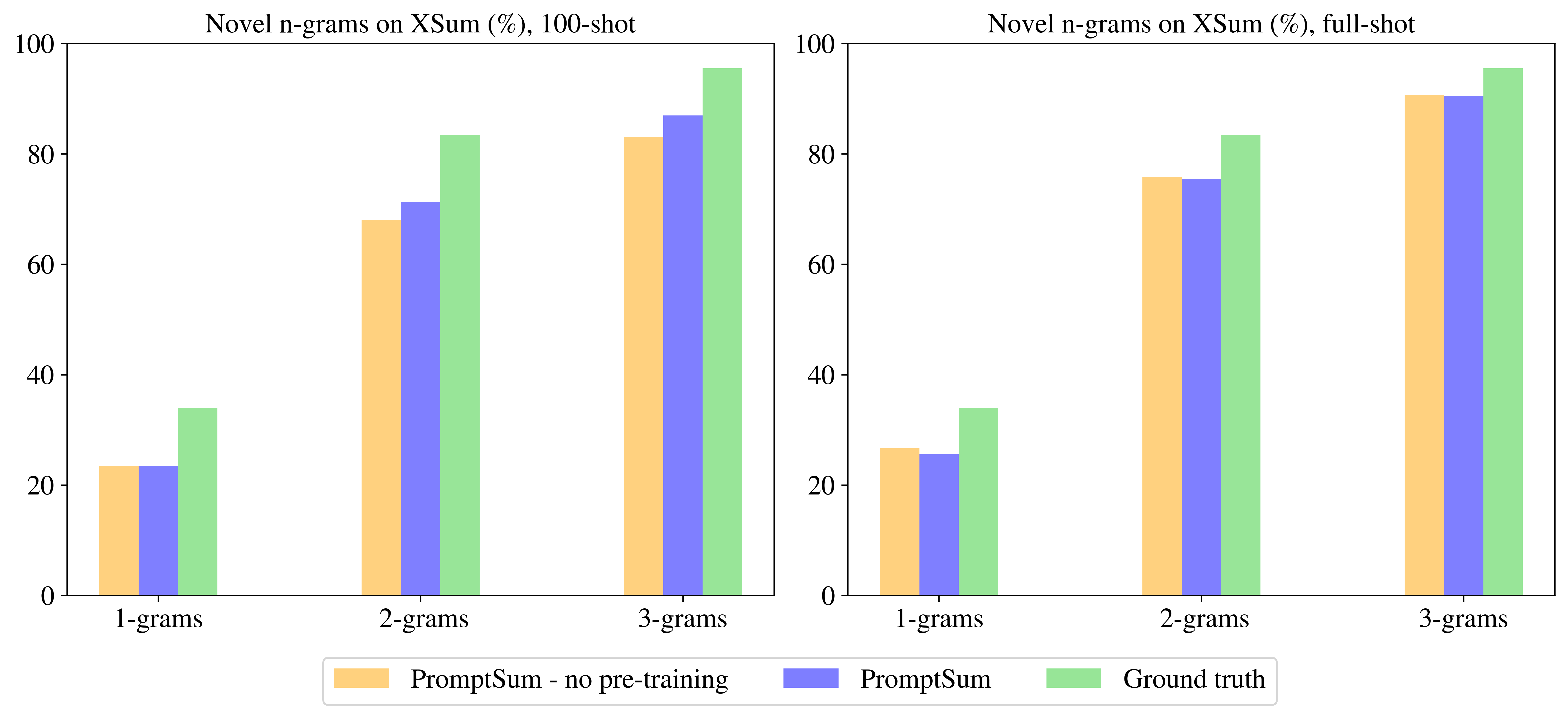}
    \includegraphics[width=\columnwidth]{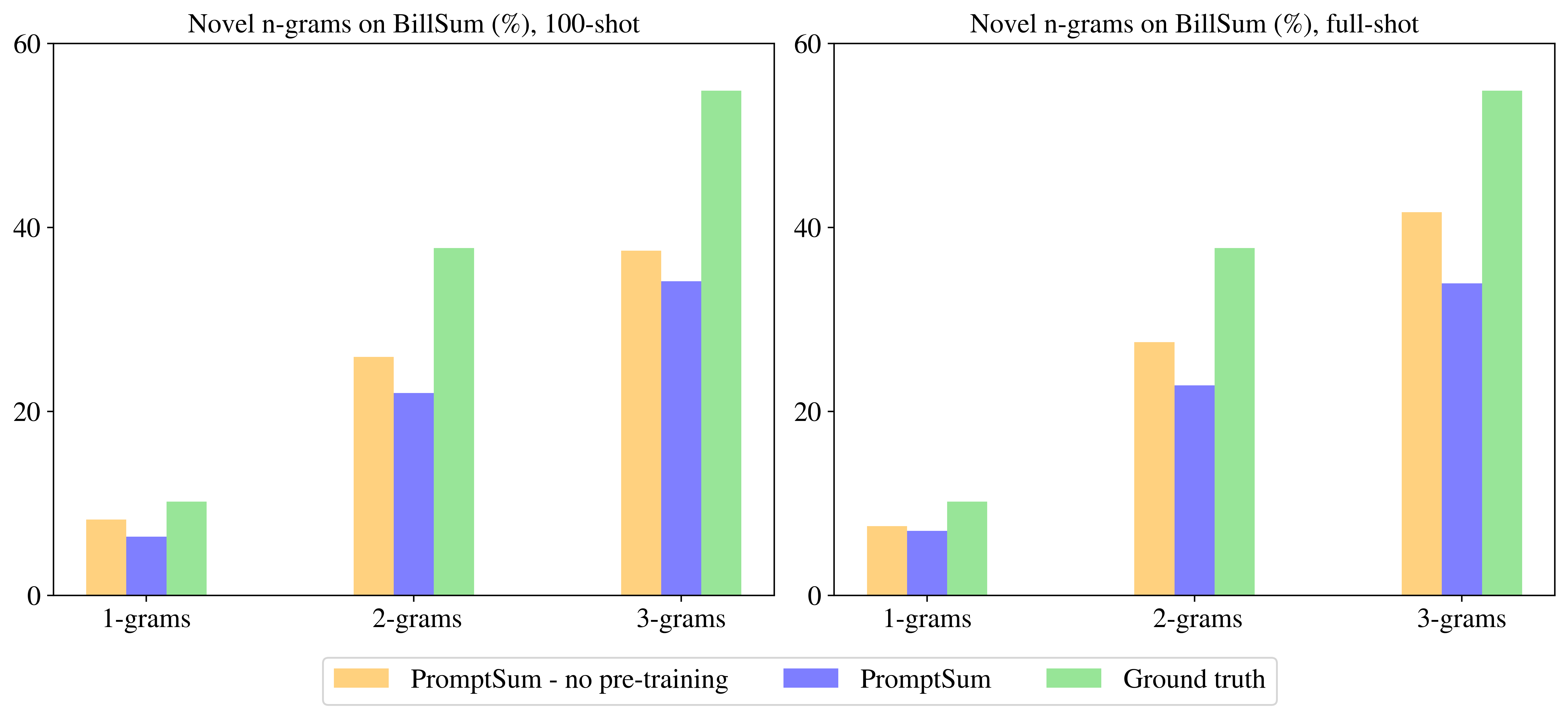}
    \includegraphics[width=\columnwidth]{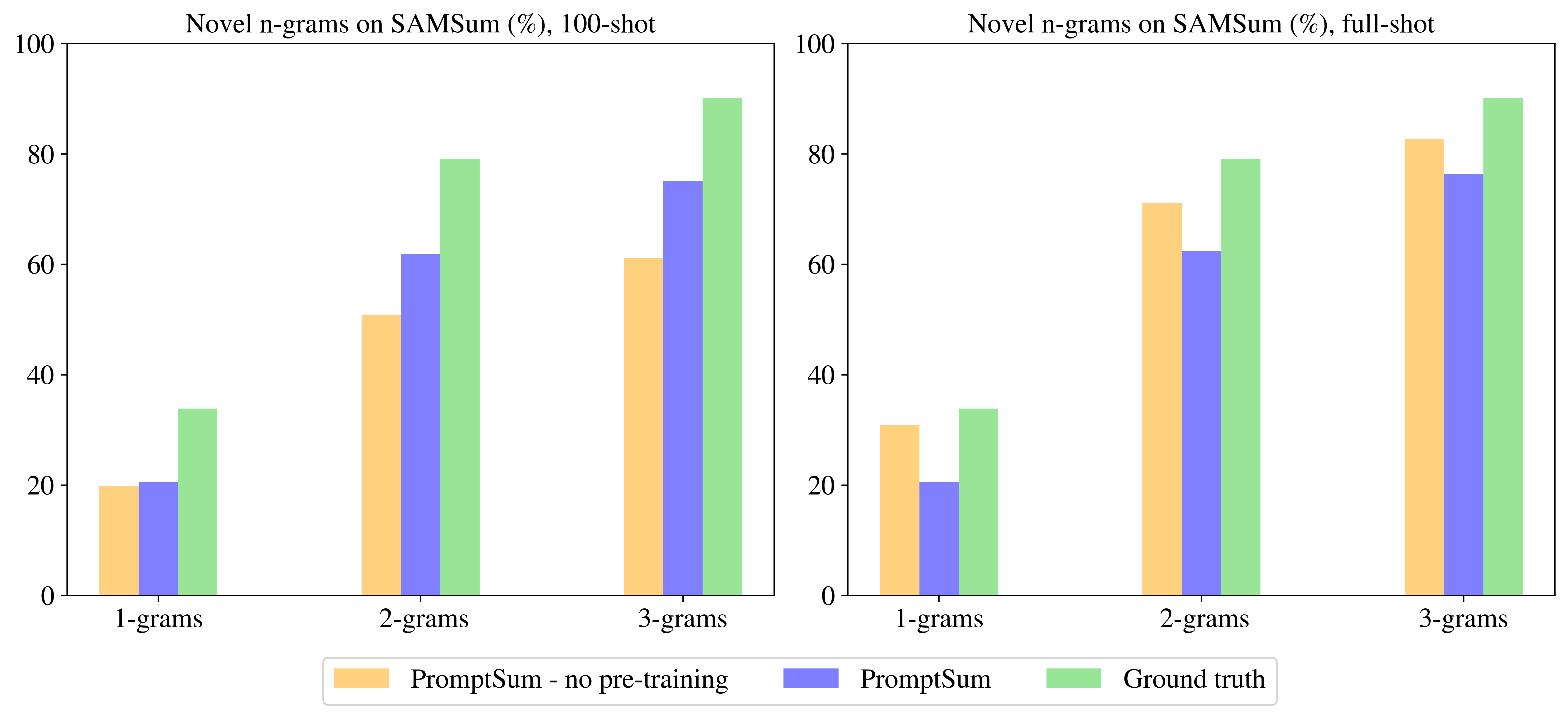}
    \caption{\small Abstractiveness levels (new n-grams percentage in generated summaries) on each of the four dataset, for 100-shot models (left) and full-shot models (right).}
    \label{fig_c}
\end{figure}

In \Cref{fig_c}, we plot abstractiveness in generated summaries for PromptSum (with and without pre-training), alongside ground truth levels. We measure abstractiveness through the fraction of unigrams, bigrams and trigrams present in the summary but not in the source. XSum being a very abstractive summarization task by design, it is difficult to isolate patterns on this dataset. On other datasets, PromptSum produces summaries less abstractive than ground truth summaries, especially on CNN/DM full-shot, a pattern which has been reported with standard models like PEGASUS and BART in the literature \citep{goyal-etal-2022-training}. In 100-shot on CNN/DM, PromptSum without pre-training shows high abstractiveness, comparable to ground-truth levels, but given the low {\sc{Rouge}} (\Cref{tab:2}), we attribute it to hallucinations.

\section{Generation Diversity}
\label{sec:appendix_d}
As seen in \Cref{tab:7}, different entity chains give very different output summaries, enabling a diversity mechanism in summary generation analogous to generation methods such as diverse beam search \cite{vijayakumar2016diverse} or top-k sampling \cite{fan2018hierarchical}. To quantify this diversity, we sample 500 data points from XSum test set, and build 10 random entity chains by sampling from source entities. We compute the \emph{oracle} {\sc{Rouge-1}} (\textbf{Oracle R-1}), corresponding to the maximum {\sc{Rouge-1}} achieved by any summary in the generated set, and the mean {\sc{Rouge}} between all pairs of summaries (\textbf{Inter R-1}). We also report the {\sc{Rouge-1}} achieved by one summary candidate sampled randomly (\textbf{Random R-1}). 

In parallel to the entity sampling describe above, we also generate 10 summary candidates with diverse beam search (using 10 groups and diversity penalty 1.0), an alternative decoding method to beam search designed to produce more varied generated outputs. Lastly, we combine both entity sampling and diverse beam search as diversity generation methods, leading to 100 summary candidates (10 diverse beams for each of the 10 entity chains) per data point.

\begin{table}[h]
\resizebox{\columnwidth}{!}{
\begin{tabular}{llcccc}

\toprule 

\textbf{Model}                    
& \textbf{Candidate generation}
& \textbf{\# Candidates} 
& \textbf{Random R-1} ($\uparrow$)
& \textbf{Oracle R-1} ($\uparrow$)
& \textbf{Inter R-1} ($\downarrow$) \\

\midrule 

PEGASUS*                  
& DBS                     & 10 & 38.13 & 50.86 & 52.58 \\



\cdashline{1-6}

\multirow{3}{*}{\begin{tabular}[c]{@{}l@{}}PromptSum \\ (100-shot)\end{tabular}} 
& DBS                     & 10 & \textbf{38.50} & 49.55 & 56.51 \\
& Entity sampling                & 10 & 35.04 & 47.38 & 49.53 \\
& DBS + Entity sampling          & 100 & 34.20 & \textbf{54.94} & \textbf{42.39} \\

\bottomrule

\end{tabular}
}
\caption{\small Diversity analysis. We generate multiple candidates on 500 test points of XSum either through diverse beam search (\textbf{DBS}) or by entity sampling from source entities. *PEGASUS by itself is not controllable, preventing an entity sampling mechanism inducing candidates diversity, thus we stick to diverse beam search only for PEGASUS.}
\label{tab:d}
\end{table}

Results of this experiment are in \Cref{tab:d}. Both diverse beam search and entity sampling show high variance in output summaries, which clinch Inter and Oracle {\sc{Rouge-1}} of 49.53 and 47.38 respectively, much above random candidates and results reported in \Cref{tab:2}. This proves that selecting a good entity chain can yield extremely high summary output performance. Moreover, \textbf{Inter R-1} is significantly lower with entity sampling than diverse beam search, suggesting that entity sampling produces more diverse summaries, which overlap less with each other. These patterns emerge even more strongly when combining entity sampling with diverse beam search.

\section{Controllability Examples}
\label{sec:appendix_e}
In the following Tables, we show controllability examples following the same format as \Cref{tab:7} on the other datasets: CNN/DM (\Cref{tab:e_1}), BillSum (\Cref{tab:e_2}), and SAMSum (\Cref{tab:e_3}), all with PromptSum trained in 100-shot.

\begin{table*}[]
\centering
\resizebox{\textwidth}{!}{
\begin{tabular}{p{4cm}p{14cm}}
\toprule
\textbf{Entity Chain} 
& \textbf{Summary} \\

\midrule

\_ & \small \textbf{Source} A man who launched an online campaign to help him destroy his brand new \$49,000 Jeep will read out a 'not sorry' apology to the company on national television tonight. In June last year, Ashton Wood raised \$18,000 online so he and 300 people could destroy his car after Jeep refused to pay a full refund for the car or replace the vehicle which he claimed had suffered 21 separate mechanical problems. During a failed settlement between the manufacturer, Fiat Chrysler Automobiles' Jeep, and Mr Wood, one of the requests made by the car company was for the disgruntled buyer to publish an 'apology' in a national publication. However, because no one would publish the apology, he will instead read it out on the ABC's 'The Checkout' tonight. Scroll down for video. Ashton Wood issued this apology to Jeep after the car company requested he apologise for criticising them. The Queenslander launched a public campaign last year asking others to help him destroy his car in protest. Compiled by Mr Wood and the ABC's The Checkout, the apology reads: 'My personal apology to Jeep: I'm Sorry. 'Dear Jeep, I'm sorry. I'm sorry my Jeep broke down before it even left the dealership. 'I'm sorry it had 21 problems. I'm sorry it had to be towed four times. I'm sorry I wasn't more grateful for your offer of a new battery. 'I'm sorry you wouldn't offer me a replacement or refund. But most of all, I'm sorry I bought a Jeep.' Mr Wood launched a Kickstarter campaign last year in an effort to raise \$22,500 - the amount Chrysler offered him to buyback the car - and allow donors the chance to be involved in the car's destruction. He is now campaigning for the United State's Lemon laws to be introduced in Australia. Pictured: 'lemon car' drivers protest against vehicle companies who refuse to refund or replace their cars. Mr Wood's extreme measures came after years of frustration with his 'cursed' Jeep Cherokee, which in the four years since he bought it in 2010, had numerous problems. [...] \\

\cdashline{1-2}

\_ & \small \textbf{Ground Truth Summary} Ashton Wood has issued a tongue-in-cheek apology to \textcolor{Brown}{Jeep} and will read it out on \textcolor{RawSienna}{ABC} TV's The Checkout tonight. The car company requested he apologise for slamming them after he was not given a refund or replacement for a faulty vehicle he bought in \textcolor{YellowGreen}{2010}. Mr \textcolor{Bittersweet}{Wood} instead launched an online campaign to destroy his \$\textcolor{Magenta}{49,000} car. He is now campaigning to introduce Lemon Laws in \textcolor{NavyBlue}{Australia} A spokesperson for \textcolor{Brown}{Jeep} said 'we have and always will treat Mr \textcolor{Bittersweet}{Wood} in a fair and professional manner. \\

\cdashline{1-2}

\small \textcolor{NavyBlue}{Australia} 
& \small A man who launched an online campaign to help him destroy his brand new \$49,000 Jeep will read out a 'not sorry' apology to the company on national television. He is now campaigning for the United State's Lemon laws to be introduced in \textcolor{NavyBlue}{Australia}. He claims years of frustration with his 'cursed' Jeep Cherokee, which in the four years since he bought it in 2010, had numerous problems. However, because no one would publish the apology, he will instead read it out on the ABC's 'The Checkout' tonight. \\

\cdashline{1-2}

\small \textcolor{ForestGreen}{June last year}
& \small A man who launched an online campaign to help him destroy his brand new \$49,000 Jeep will read out a 'not sorry' apology to the company on national television. In \textcolor{ForestGreen}{June last year}, Mr Wood raised \$18,000 online so he and 300 people could destroy his car after Jeep refused to pay a full refund for the car or replace the vehicle which he claimed had suffered 21 separate mechanical problems. However, because no one would publish the apology, he will instead read it out on the ABC's 'The Checkout' tonight. \\

\cdashline{1-2}

\small \textcolor{RubineRed}{21}, \textcolor{BlueGreen}{The United States} 
& \small A man who launched an online campaign to help him destroy his brand new \$49,000 Jeep will read out a 'not sorry' apology to the car company on national television. He claims \textcolor{RubineRed}{21} separate mechanical problems with his car. He is campaigning for \textcolor{BlueGreen}{The United States} Lemon laws to be introduced in Australia. However, because no one would publish the apology, he will instead read it out on the ABC's 'The Checkout' tonight. \\

\cdashline{1-2}

\small \textcolor{ForestGreen}{June last year}, \textcolor{YellowGreen}{2010} 
& \small A man who launched an online campaign to help him destroy his brand new \$49,000 Jeep will read out a 'not sorry' apology to the company on national television. In \textcolor{ForestGreen}{June last year}, Mr Wood raised \$18,000 online so he and 300 people could destroy his car after Jeep refused to pay a full refund for the car or replace the vehicle which he claimed had suffered 21 separate mechanical problems. He claims years of frustration with his 'cursed' Jeep Cherokee, which in the four years since he bought it in \textcolor{YellowGreen}{2010}, had numerous problems. \\

\cdashline{1-2}

\small \textcolor{Brown}{Jeep}, \textcolor{Orchid}{300}, \textcolor{Magenta}{49,000} 
& \small A man who launched an online campaign to help him destroy his brand new \$\textcolor{Magenta}{49,000} \textcolor{Brown}{Jeep} will read out a 'not sorry' apology to the car company on national television. However, because no one would publish the apology, he will instead read it out on the ABC's 'The Checkout'. Pictured: Mr Wood and \textcolor{Orchid}{300} people destroying his brand new \$\textcolor{Magenta}{49,000} car. \\

\cdashline{1-2}

\small \textcolor{RawSienna}{ABC}, \textcolor{NavyBlue}{Australia}, \textcolor{JungleGreen}{the four years} 
& \small A man who launched an online campaign to help him destroy his brand new \$49,000 Jeep will read out a 'not sorry' apology to the company on national television. However, because no one would publish the apology, he will instead read it out on the \textcolor{RawSienna}{ABC}'s 'The Checkout' tonight. He is now campaigning for the United State's Lemon laws to be introduced in \textcolor{NavyBlue}{Australia}. He claims years of frustration with his 'cursed' Jeep Cherokee, which in \textcolor{JungleGreen}{the four years}  since he bought it in 2010, had numerous problems. \\

\cdashline{1-2}

\small \textcolor{BlueGreen}{The United States}, \textcolor{Sepia}{Kickstarter}, \textcolor{Bittersweet}{Wood} 
& \small A man who launched an online campaign to help him destroy his brand new \$49,000 Jeep will read out a 'not sorry' apology to the company on national television. He is now campaigning for \textcolor{BlueGreen}{the United States} Lemon laws to be introduced in Australia. \textcolor{Sepia}{Kickstarter} campaign helped Mr \textcolor{Bittersweet}{Wood}  raise \$22,500 - the amount Chrysler offered him to buyback the car. However, because no one would publish the apology, he will instead read it out on the ABC's 'The Checkout' tonight. \\

\bottomrule

\end{tabular}
}
\caption{\small Controllability qualitative examples on CNN/DM 100-shot. Entities are sampled from source entities.}
\label{tab:e_1}
\end{table*}

\begin{table*}[]
\centering
\resizebox{\textwidth}{!}{
\begin{tabular}{p{4cm}p{14cm}}
\toprule
\textbf{Entity Chain} 
& \textbf{Summary} \\

\midrule

\_ & \small \textbf{Source} INPUT: SECTION 1. SHORT TITLE. This Act may be cited as the Coastal and Estuarine Land Protection Act''. SEC. 2. FINDINGS. Congress finds the following: (1) Coastal and estuarine areas provide important nursery habitat for two-thirds of the United States commercial fish and shellfish, provide nesting and foraging habitat for coastal birds, harbor significant natural plant communities, and serve to facilitate coastal flood control and pollutant filtration. (2) The Coastal Zone Management Act of 1972 (16 U.S.C. 1451 et seq.) recognizes the national importance of these areas and their ecological vulnerability to anthropogenic activities by establishing a comprehensive Federal and State partnership for protecting natural reserves and managing growth in these areas. (3) The National Estuarine Research Reserve system established under that Act relies on the protection of pristine designated areas for long-term protection and for the conduct of education and research critical to the protection and conservation of coastal and estuarine resources. (4) Intense development pressures within the coastal watershed are driving the need to provide coastal managers with a wider range of tools to protect and conserve important coastal and estuarine areas. (5) Protection of undeveloped coastal lands through the acquisition of interests in property from a willing seller are a cost-effective means of providing these areas with permanent protection from development. (6) Permanent protection of lands in the coastal zone is a necessary component of any program to maintain and enhance coastal and estuarine areas for the benefit of the United States, including protection of water quality, access to public beachfront, conserving wildlife habitat, and sustaining sport and commercial fisheries. (7) Federal, State, and nongovernmental organization pilot land acquisition projects have already substantially contributed to the long-term health and viability of coastal and estuarine systems. (8) Enhanced protection of estuarine and coastal areas can be attained through watershed-based acquisition strategies coordinated through Federal, State, regional, and local efforts. (9) Conserving coastal and estuarine lands can support the traditional economic and natural resource bases of communities in the coastal watershed, including well-managed forests that demonstrate outstanding ecological, recreational, historical, and aesthetic attributes. SEC. 3. ESTABLISHMENT OF COASTAL AND ESTUARINE LAND PROTECTION PROGRAM. (a) In General.-- (1) Establishment.--The Secretary of Commerce shall establish a Coastal and Estuarine Land Protection Program (hereinafter referred to as the program''), in cooperation with appropriate State, regional, and other units of Government for the purposes of protecting the environmental integrity of important coastal and estuarine areas, [...] \\

\cdashline{1-2}

\_ & \small \textbf{Ground Truth Summary} Coastal and Estuarine Land Protection Act - Directs the Secretary of \textcolor{Maroon}{Commerce}  to establish a Coastal and \textcolor{Bittersweet}{Estuarine Land Protection Program} to protect the environmental integrity of undeveloped coastal and estuarine areas. Requires the program to be administered by the National Ocean Service of the National Oceanic and Atmospheric Administration (NOAA) through the Office of Ocean and Coastal Resource Management. Authorizes the Secretary to make Program grants to coastal states with approved coastal zone management plans or National Estuarine Research Reserve units for the purpose of acquiring property that will further the goals of an approved Coastal Zone Management Plan or Program, a National Estuarine Research Reserve management plan, or a regional or state watershed protection plan. Prohibits any more than 75\% of the funding for any project funded with a grant made under the program from being derived from federal sources. Reserves 15\% of program funds for acquisitions benefitting the National Estuarine Research Reserve. Specifies that when property is acquired under this program, the grant recipient shall provide assurances that: (1) title will be held by the recipient or another public agency designated by the recipient in perpetuity; (2) property will be managed consistent with the purpose of the Program; and (3) funds will be returned to the Secretary for redistribution if the property is sold, exchanged, or divested. \\

\cdashline{1-2}

\small \textcolor{Magenta}{two-thirds}, \textcolor{Brown}{The Coastal Zone Management Act} 
& \small Coastal and estuarine Land Protection Act - Amends \textcolor{Brown}{The Coastal Zone Management Act} to provide for permanent protection of undeveloped coastal lands through the acquisition of interests in property from a willing seller. Coastal and estuarine areas provide important nursery habitat for \textcolor{Magenta}{two-thirds} of the United States commercial fish and shellfish, provide nesting and foraging habitat for coastal birds, harbor significant natural plant communities, and serve to facilitate coastal flood control and pollutant filtration. \textcolor{Brown}{The Coastal Zone Management Act} recognizes the national importance of these areas and their ecological vulnerability to anthropogenic activities by establishing a comprehensive Federal and State partnership for protecting natural reserves and managing growth in these areas.
 \\

 \cdashline{1-2}

\small \textcolor{Bittersweet}{Estuarine Land Protection Program}, \textcolor{YellowGreen}{1972} 
& \small Coastal and \textcolor{Bittersweet}{Estuarine Land Protection Program} - Amends the Coastal Zone Management Act of \textcolor{YellowGreen}{1972} to establish a Coastal and \textcolor{Bittersweet}{Estuarine Land Protection Program} (hereinafter referred to as the program''), in cooperation with appropriate State, regional, and other units of Government for the purposes of protecting the environmental integrity of important coastal and estuarine areas. The purpose of the program is to provide Federal, State, regional, and local agencies with a wide range of tools to protect and conserve important coastal and estuarine areas, including protection of undeveloped coastal lands through the acquisition of interests in property from a willing seller. \\

 \cdashline{1-2}

\small \textcolor{NavyBlue}{the United States}, \textcolor{Magenta}{two-thirds}, \textcolor{Maroon}{Commerce} 
& \small Coastal and estuarine Land Protection Act - Establishes a program to maintain and enhance coastal and estuarine areas for the benefit of \textcolor{NavyBlue}{the United States}, including protection of water quality, access to public beachfront, wildlife habitat, and sustaining sport and commercial fisheries. Coastal and estuarine areas provide important nursery habitat for \textcolor{Magenta}{two-thirds} of \textcolor{NavyBlue}{the United States} commercial fish and shellfish, provide nesting and foraging habitat for coastal birds, harbor significant natural plant communities, and serve to facilitate coastal flood control and pollutant filtration. Directs the Secretary of \textcolor{Maroon}{Commerce} to establish a program to maintain and enhance coastal and estuarine areas for the benefit of the United States, including protection of water quality, access to public beachfront, wildlife habitat, and sustaining sport and commercial fisheries. \\

 \cdashline{1-2}

\small \textcolor{Bittersweet}{Estuarine Land Protection Program}, \textcolor{Maroon}{Commerce}, \textcolor{YellowGreen}{1972} 
& \small Coastal and Estuarine Land Protection Act - Amends the Coastal Zone Management Act of \textcolor{YellowGreen}{1972} to establish a Coastal and \textcolor{Bittersweet}{Estuarine Land Protection Program} (hereinafter referred to as the program''), in cooperation with appropriate State, regional, and other units of Government for the purposes of conserving the environmental integrity of important coastal and estuarine areas. This bill authorizes the Secretary of \textcolor{Maroon}{Commerce} to establish a Coastal and \textcolor{Bittersweet}{Estuarine Land Protection Program} (hereinafter referred to as the program''), in cooperation with appropriate State, regional, and other units of Government for the purposes of conserving the environmental integrity of important coastal and estuarine areas.\\

\bottomrule

\end{tabular}
}
\caption{\small Controllability qualitative examples on BillSum 100-shot. Entities are sampled from source entities.}
\label{tab:e_2}
\end{table*}

\begin{table*}[]
\centering
\resizebox{\textwidth}{!}{
\begin{tabular}{p{4cm}p{14cm}}
\toprule
\textbf{Entity Chain} 
& \textbf{Summary} \\

\midrule

\_ & \small \makecell[l]{\textbf{Source} \\
Matt: have you heard that Bon Jovi are coming to Poland? \\ 
Phil: No way! \\
Phil: where, when? \\
Matt: Warsaw, next July \\
Tony: would you like to go? \\
Matt: Sure! \\
Phil: how much are the tickets? Very expensive? \\
Phil: they're huge, I guess tix can cost a bomb \\
Matt: I read the cheapest ones start from pln 250 \\
Phil: not too bad, for Bon Jovi \\
Matt: Yep! \\
Tony: maybe I will go too \\
Tony: Would you like to get the tix soon? \\
Matt: Yes \\
Matt: The sooner, the better \\
Tony: Right, they may be sold out pretty soon \\
Matt: How about you, Phil? \\
Matt: would you like to go too? \\
Phil: Will think about it \\
Phil: I liked them a lot when I was a kid \\
Phil: It would be great to see them live \\
Matt: I guess so! \\
Matt: Think about it and let me know soon, ok? \\
Phil: Sure} \\

\cdashline{1-2}

\_ & \small \textbf{Ground Truth Summary} \textcolor{Bittersweet}{Matt} and \textcolor{Brown}{Tony}  want to go to the concert of \textcolor{Sepia}{Bon Jovi} \textcolor{ForestGreen}{next July}  in Poland and are planning to buy the tickets that cost \textcolor{Magenta}{250}  PLN before they’re sold out. Phil will think about it and let them know. \\

\cdashline{1-2}

\small \textcolor{NavyBlue}{Warsaw} 
& \small Bon Jovi are coming to \textcolor{NavyBlue}{Warsaw}  next July. \\

\cdashline{1-2}

\small \textcolor{Bittersweet}{Matt} 
& \small \textcolor{Bittersweet}{Matt} has heard that Bon Jovi are coming to Warsaw. \\

\cdashline{1-2}

\small \textcolor{Brown}{Tony} 
& \small Matt and \textcolor{Brown}{Tony} would like to go too. \\

\cdashline{1-2}

\small \textcolor{NavyBlue}{Warsaw}, \textcolor{Magenta}{250} 
& \small Bon Jovi are coming to \textcolor{NavyBlue}{Warsaw} and tickets start from pln \textcolor{Magenta}{250}. \\

\cdashline{1-2}

\small \textcolor{Bittersweet}{Matt}, \textcolor{ForestGreen}{next July} 
& \small \textcolor{Bittersweet}{Matt} has heard that Bon Jovi are coming to Warsaw \textcolor{ForestGreen}{next July}. \\

\cdashline{1-2}

\small \textcolor{Sepia}{Bon Jovi}, \textcolor{Brown}{Tony} 
& \small \textcolor{Sepia}{Bon Jovi} are coming to the country and 
\textcolor{Brown}{Tony} would like to go too. \\

\cdashline{1-2}

\small \textcolor{Brown}{Tony}, \textcolor{NavyBlue}{Warsaw}, \textcolor{Sepia}{Bon Jovi} 
& \small \textcolor{Brown}{Tony} would like to go to \textcolor{NavyBlue}{Warsaw} for the \textcolor{Sepia}{Bon Jovi} concert. \\

\cdashline{1-2}

\small \textcolor{ForestGreen}{next July}, \textcolor{NavyBlue}{Warsaw}, \textcolor{Magenta}{250} 
& \small Bon Jovi are coming \textcolor{ForestGreen}{next July} to \textcolor{NavyBlue}{Warsaw}. The cheapest tickets start from pln \textcolor{Magenta}{250}. \\

\bottomrule

\end{tabular}
}
\caption{\small Controllability qualitative examples on SAMSum 100-shot. Entities are sampled from source entities.}
\label{tab:e_3}
\end{table*}

\end{document}